\documentclass[11pt]{article}

\usepackage[final]{acl}

\usepackage{times}
\usepackage{latexsym}
\usepackage{adjustbox}
\usepackage{tcolorbox}
\tcbuselibrary{skins}
\tcbuselibrary{breakable}

\usepackage[T1]{fontenc}
\usepackage[utf8]{inputenc}

\usepackage{microtype}

\usepackage{inconsolata}

\usepackage{graphicx}

\usepackage{hyperref}       % hyperlinks
\usepackage{url}            % simple URL typesetting
\usepackage{booktabs}       % professional-quality tables
\usepackage{amsfonts}       % blackboard math symbols
\usepackage{nicefrac}       % compact symbols for 1/2, etc.
\usepackage{microtype}      % microtypography
\usepackage{xcolor}         % colors
\usepackage{multirow}

\usepackage{graphicx}
\usepackage{subcaption}
\usepackage{booktabs} % for professional tables
\usepackage{tcolorbox}
\usepackage{caption}
\usepackage{amsmath}
\usepackage{enumitem}

\usepackage{titlesec}

\usepackage{makecell}
\usepackage{amssymb}

\usepackage[capitalize,noabbrev]{cleveref}
\usepackage[table]{xcolor}
\newtcolorbox{rqfinding}{colback=violet!8, colframe=violet!55, arc=3pt, boxrule=0.7pt, left=5pt, right=5pt, top=4pt, bottom=4pt, fontupper=\small}
\DeclareMathOperator{\logit}{logit}

\title{\bench:\\Disentangling Knowledge Gaps in Multilingual LLM Evaluation}

\author{
\textbf{Angelika Romanou\textsuperscript{1 *}},
\textbf{María Grandury\textsuperscript{1 *}},
\textbf{Clara Meister\textsuperscript{1}},
\textbf{Negar Foroutan\textsuperscript{1}},\\
\textbf{Jabez Magomere\textsuperscript{2}},
\textbf{Elif Kurtay\textsuperscript{1}},
\textbf{Anna Sotnikova\textsuperscript{1}},
\textbf{Orestis Papakyriakopoulos\textsuperscript{3}},
\textbf{Scott A. Hale\textsuperscript{2}},\\
\textbf{Shamsuddeen Hassan Muhammad\textsuperscript{4}},
\textbf{Antoine Bosselut\textsuperscript{1}}
\\[0.5em]
\textsuperscript{1}EPFL,
\textsuperscript{2}University of Oxford,
\textsuperscript{3}Technical University of Munich,
\textsuperscript{4}Imperial College London
\\[0.5em]
\small
Akshay Kumar Dubey, Alfonso Amayuelas, Amit Agarwal, Arion Das, Carlos Arriaga, Cheng Yu, Dalia Ali,  David Quispe-Castillo, \\ \small Enrique Gabriel, Paiva Galeano, Fenal Ashokbhai Ilasariya, Francisco-Javier  Rodrigo-Ginés,  Gonzalo Martínez, Gouthami Vadithya, \\ \small Guido Ivetta, Hitesh Laxmichand Patel, Javier Conde, Jorge Vallego,  Juan J. Calderón G.,  Karan Gupta,  Marcos J. Gomez,  \\ \small Marvin M. Agüero-Torales, Mithil Bangera, Naira Paola Arnez Jordan, Oscar M. Cumbicus-Pineda, Priyaranjan Pattnayak, \\ \small Reewos Erwin Talla Chumpitaz, Sahil Aggarwal,  Sahil Raj, Shruti Jalan, Srikant Panda, Susana Zhou, Yeshil Bangera
\normalsize
}

\begin{document}

\newcommand{\bench}{\texttt{INCLUDE-2.0}}
\newcommand{\benchvone}{\texttt{INCLUDE}}
\newcommand{\eg}{\textit{e.g.}}
\newcommand{\ie}{\textit{i.e.}}
\newcommand{\fixme}{{\color{red}FIX ME}}

\definecolor{blue_back}{RGB}{207,226,243}
\definecolor{green_back}{RGB}{217,234,211}
\definecolor{yellow_back}{RGB}{255,242,204}
\definecolor{orange_back}{RGB}{252,229,205}
\definecolor{purple_back}{RGB}{217,210,233}
\definecolor{blue_back}{HTML}{f1f5f8}

\definecolor{saffron}{RGB}{227, 170, 0}
\definecolor{pink}{RGB}{204, 0, 102}
\definecolor{asparagus}{rgb}{0.53, 0.66, 0.42}
\definecolor{forestgreen(web)}{rgb}{0.13, 0.55, 0.13}
\definecolor{purple2}{RGB}{174,159,208}
\newcommand{\AR}[1]{\textcolor{pink}{\textit{(Angie: #1)}}}
\newcommand{\nf}[1]{\textcolor{forestgreen(web)}{\textit{(Negar: #1)}}}
\newcommand{\cm}[1]{\textcolor{orange}{\textit{(Clara: #1)}}}
\newcommand{\jm}[1]{\textcolor{teal}{\textit{(Jabez: #1)}}}
\newcommand{\mg}[1]{\textcolor{purple2}{\textit{(María: #1)}}}

\maketitle
\begingroup
\renewcommand{\thefootnote}{*}
\footnotetext{Equal contribution.}
\endgroup

\begin{abstract}

Current multilingual benchmarks cover %only a fraction of the languages represented in modern pretraining corpora and 
often conflate language proficiency with the ability to answer questions grounded in specific regions and communities. 
We introduce \bench{}, a large-scale benchmark for region-aware multilingual factual evaluation, which covers 91 languages from 86 countries (127 total language--country pairs), comprising over 220K curated samples drawn from native regional sources. 
Unlike existing benchmarks that often treat language as a proxy for region, \bench{} distinguishes between questions whose answers are independent of context, vary by country, concern widely documented facts about a place, or depend on knowledge primarily expressed through local linguistic and cultural sources. 
Each item is annotated with its regional knowledge type and domain, and paired with an English translation, enabling controlled analysis of the relationship between language proficiency and possession of regional knowledge. Evaluating frontier and open-weight models on \bench{}, we find that aggregate measures of multilingual accuracy mask substantial performance differenes across regional knowledge types, languages, and regions, and that frontier models generally perform worse on locally embedded knowledge than other knowledge types. %on widely documented facts about regions., with an average drop of $\sim$14.5 points. We also show that for open-weight models, scale improves accuracy but does not reliably make performance invariant to whether a question is asked in English or its original language.
\bench{} is available at \url{https://huggingface.co/datasets/epfl-nlp/include-2.0}

\end{abstract}

\begin{figure}[t]
  \centering
  \includegraphics[width=\columnwidth]{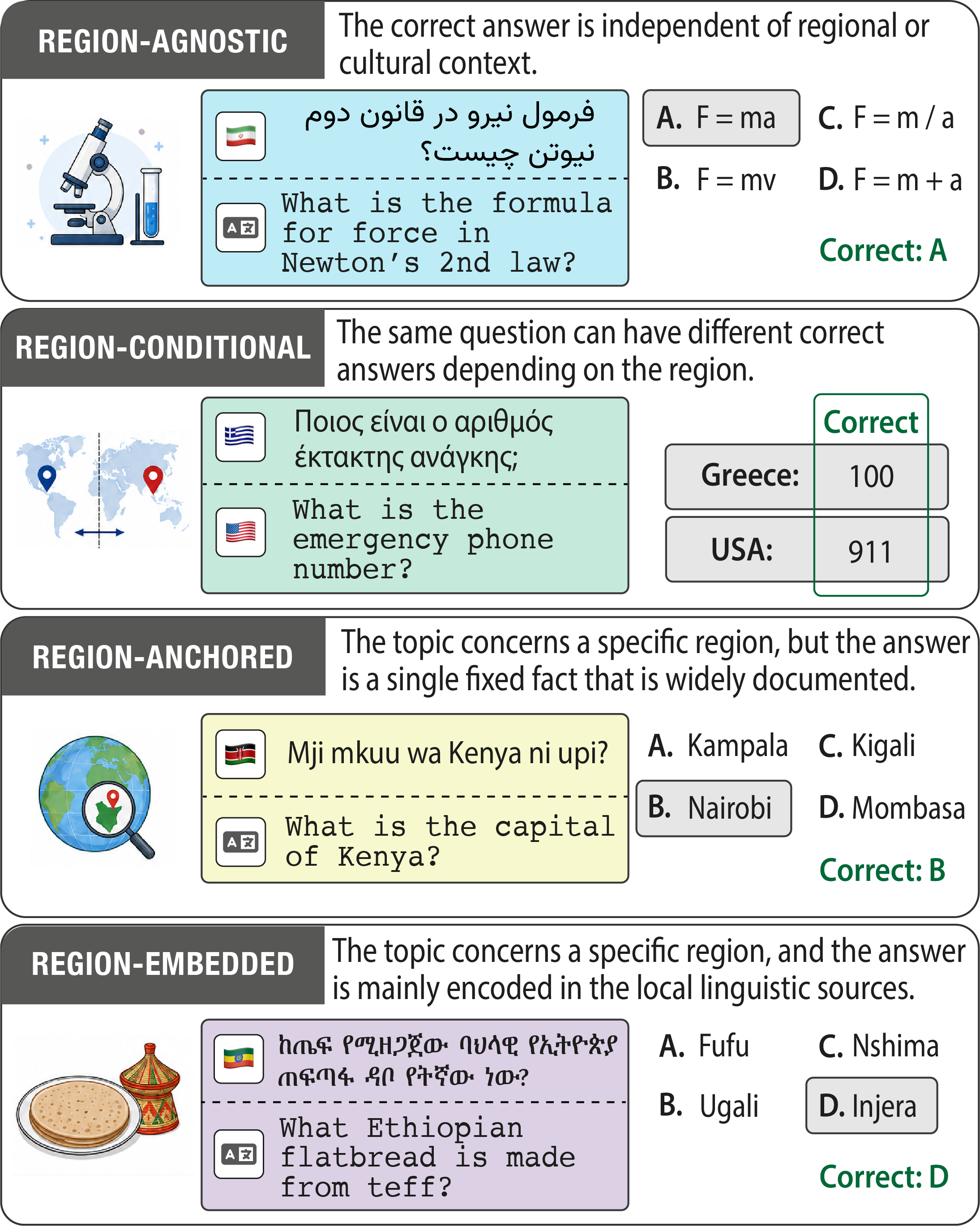}
  % \caption{Regional factual knowledge taxonomy in \bench{}. 
  % The taxonomy separates questions whose answers are independent of regional context, 
  % depend on the specified region, concern widely documented facts about a region, 
  % % or require knowledge primarily available through local linguistic and cultural sources.
  % }
\caption{Regional factual knowledge taxonomy in \bench{}.
The taxonomy separates questions whose answers are region-independent,
region-dependent, widely documented regional facts, or available through local linguistic and cultural sources.
}
  \label{fig:main}
\end{figure}

\begin{table*}[h]
\centering
\resizebox{\textwidth}{!}{
\begin{tabular}{l|ccccccc}
\toprule
\textbf{Benchmark} & \textbf{\#Languages} & \textbf{\#Scripts} & \textbf{\#Countries} & \textbf{\makecell{\#Language-- \\ Country Pairs}} & \textbf{Knowledge Type} & \textbf{\makecell{Region \\ Bias Analysis}} & \textbf{Data Origin} \\
\midrule
\texttt{GlobalMMLU} & 42 & 13 & 1 & -- & US-centric & $\checkmark$ & Human/Machine Translated \\
\texttt{MMMLU} & 14 & 8 & 1 & -- & US-centric & -- & Machine-Translated \\
\texttt{MMLU-ProX} & 29 & 11 & 1 & -- & US-centric & -- & Human/Machine Translated \\
\texttt{EXAMS} & 16 & 3 & -- & -- & Regional \& Global & -- & \textbf{Native} \\
\texttt{MultiLoKo} & 31 & 12 & -- & -- & Regional & -- & \textbf{Native} + Human/Machine Translated \\
\texttt{BLEND} & 13 & 7 & 16 & 16 & Regional & -- & Native + Machine Translated \\
\texttt{CulturalBench} & 1 & 1 & 45 & 45 & Regional & -- & Human + Synthetic Generation \\
\texttt{INCLUDE} & 44 & 15 & 52 & -- & Regional \& Global & -- & \textbf{Native} \\
\midrule
\rowcolor{blue_back}
\bench{} (Ours) & \textbf{91} & \textbf{22} & \textbf{86} & \textbf{127} & Regional \& Global & $\checkmark$ & \textbf{Native} \\

\bottomrule
\end{tabular}
}
\caption{Benchmark comparison. 
\bench{} offers the broadest coverage across languages, scripts, and countries, while uniquely combining regional and global knowledge, region-bias analysis, and native data collection.}
\label{tab:benchmark-comparison}
\end{table*}

\section{Introduction}

% \AR{Intro is WIP}

Large language models (LLMs) are increasingly deployed across languages, countries, and cultural contexts, making it important to evaluate whether they encode the knowledge needed to serve users in the regions where those languages are spoken. 
Current multilingual evaluations, however, often conflate language proficiency with regional grounding, particularly when translated from an initially-English dataset. 
A model may fail because it does not understand the language, because it lacks the relevant regional facts, or because it defaults to English- or Western-centric assumptions when faced with locally grounded questions. 
As illustrated in Figure~\ref{fig:main}, factual questions may differ substantially in how regional their answers are: some are independent of region, some require a regional context to determine the answer, some concern widely documented facts about a region, and others depend on locally embedded linguistic or cultural knowledge. 
Conflating these cases precludes diagnosing whether multilingual models genuinely support diverse regional contexts, or merely express globally widespread knowledge in many languages.

Recent benchmarks address this problem by moving beyond translated English datasets. 
Benchmarks such as \benchvone{}~\citep{romanou2024include}, EXAMS~\citep{hardalov-etal-2020-exams}, BLEnD~\citep{blend}, CulturalBench~\citep{chiu-etal-2025-culturalbench}, and Kaleidoscope~\citep{salazar2025kaleidoscope} introduce questions grounded in local educational, cultural, and regional contexts. 
However, important gaps remain. 
First, language coverage is still narrow relative to the number of languages present in modern pretraining corpora. 
As shown in Table~\ref{tab:benchmark-comparison}, prior multilingual factual and regional benchmarks cover at most 44 languages, a small fraction of the 1000+ languages represented in modern pretraining corpora \cite{penedo2025fineweb2}. 
%, whereas \bench{} covers 91 languages across 22 scripts and 86 countries.
Second, many evaluations treat language as a proxy for region, even though the same language may be used across countries with different laws, institutions, political \& social systems, and cultural practices. 
Third, regional knowledge is often treated as a single umbrella category, mixing questions that test general language competence, globally documented facts about a region, and knowledge that is deeply embedded in local context. 
As a result, existing evaluations conflate different failure modes, such as language understanding, factual recall, or region-specific grounding.

To address these limitations, we introduce \bench{}, a multilingual benchmark for region-aware knowledge evaluation. 
\bench{} builds on \benchvone{} \citep{romanou2024include}, but substantially extends it in both coverage and diagnostic structure, expanding coverage from 44 to 91 languages, covering 127 language--country pairs. Beyond scale, \bench{} introduces the regional taxonomy shown in Figure~\ref{fig:main}: \textit{region-agnostic}, \textit{region-conditional}, \textit{region-anchored}, and \textit{region-embedded}, disentangling linguistic competence from regional factual grounding. 
% English translations for every item in the dataset enable controlled comparisons between in-language and English prompting on the same underlying questions.

We summarize our contributions as threefold:

\begin{itemize}[noitemsep, topsep=2pt]
    \item \textbf{Benchmark resource.} 
    We introduce \bench{}, covering 91 languages, 22 scripts, 86 countries, 127 language--country pairs, and over 220,000 samples. 
    \bench{} specifically includes the largest factual knowledge collection for African languages to date, with 32 African languages, and one of the most diverse Indian language coverages, with 14 languages. 
    Each item is annotated with its regional knowledge category, and all items are paired with English translations for controlled analysis.
    \item \textbf{Conceptual framework.} 
    We propose a taxonomy of regional knowledge that operationalizes regionality and enables multilingual performance to be decomposed into language ability and regional knowledge.
    \item \textbf{Analyses.} 
    Using \bench{}, we analyze current LLMs' understanding of multilingual factual knowledge, whether multilingual competence improves with model scale, how language ability interacts with regional knowledge, and whether models exhibit regional biases within pluricentric languages.
\end{itemize}

\section{Operationalizing Regional Knowledge}
\label{sec:regional_operationalization}

% Regional knowledge in the context of multilingual datasets and benchmarks refers to the specific information, culture, practices, legal frameworks, and historical or social norms that are relevant to a local environment or region, and which are necessary for proper understanding and use of language in that particular context. This knowledge is often critical for language models and systems designed to operate across multiple languages and cultural settings, as it ensures that these technologies can interact effectively within the preferred linguistic and cultural backgrounds of users, countering biases introduced by overrepresentation of global (often Western) content.
% In this section, we outline three dimensions of regionality: types of regional knowledge, availability, and utilization, and discuss how to assess construct validity and operationalize each component.

In multilingual evaluation, \textit{regional knowledge} refers to information, practices, conventions, and norms whose validity, prevalence, or recognition depends on a specific geographic or cultural context \citep{romanou2024include}. 
Capturing this dimension is essential for benchmarks that are designed to measure not only linguistic competence but also a model's grounding in the linguistic contexts where it is deployed. 
Without it, evaluation conflates language proficiency with the Western-centric defaults that dominate multilingual evaluation benchmarks. 
We operationalize regionality along % three dimensions:
the \textit{type} of knowledge a question tests, and the \textit{availability} of that knowledge across linguistic communities. %, and the \textit{utilization} required to bring it to bear on the question.
% Together, these dimensions enable a decomposition of multilingual performance into linguistic and regional components.

\subsection{Regional Knowledge Types}
\label{sec:regional_taxonomy}

We define four categories based on the relationship between a question and the region it concerns. 
{
\titlespacing*{\paragraph}{0pt}{0.25ex}{1.5ex}
\paragraph{Region-agnostic.} The correct answer is independent of any regional or cultural context.
Examples include pure mathematics, physical laws, formal logic, and biochemical mechanisms.
Cross-lingual performance gaps on such knowledge are therefore more likely to reflect language-related factors rather than gaps in region-specific information.
This property makes region-agnostic knowledge a useful control axis for disentangling linguistic effects from regional or cultural factors.}
{
\titlespacing*{\paragraph}{0pt}{0.25ex}{1.5ex}
\paragraph{Region-conditional.} 
The same question can have different correct answers depending on the region.
This category includes region-specific rules and institutions, such as speed limits, voting ages, prescription policies, civil procedure, and professional licensing requirements, as well as topics whose accepted answers may vary across national curricula, public institutions, or local conventions, such as historical interpretations, religious or political positions, and disputed events. 
In all cases, the answer is fixed within a given regional context but may change across regions. 
This is especially important for languages spoken across multiple countries (\eg, Spanish, Arabic, English, French, Portuguese), where the language alone is not sufficient to determine the correct answer.
}

{
\titlespacing*{\paragraph}{0pt}{0.25ex}{1.5ex}
\paragraph{Region-anchored.} The topic of a query concerns a specific region, but the answer is a fixed globally documented fact.
Examples include major historical events, internationally known figures, or large-scale demographic information.
Such items test whether a model has acquired information \textit{about} a region, even though the information is typically accessible through widely indexed sources, including English-language reference materials.
% but the information is typically available in widely indexed sources, including English-language reference works.
}

{
\titlespacing*{\paragraph}{0pt}{0.25ex}{1.5ex}
\paragraph{Region-embedded.} The topic concerns a specific region and has a fixed answer, but the underlying knowledge is primarily encoded within a particular linguistic or cultural community and is sparsely represented elsewhere. Examples include traditional musical instruments, vernacular cuisine, folk practices, local historical figures, and regionally specialized terminology. 
Unlike region-anchored knowledge, region-embedded knowledge is typically learnable only through native-language or specialized regional sources.
% The distinction from region-anchored items is not topical but evidentiary: region-embedded knowledge is typically learnable only from native-language or specialized regional sources.
}

\subsection{Regional Knowledge Availability}
\label{sec:regional_availability}

Availability captures the extent to which a region-coupled fact is represented across cross-lingual discourse. 
We model this as a continuum operationalized by a heuristic question used during annotation: \textit{Is this fact reliably documented in major English-language reference sources, or does retrieval typically require native-language or specialized regional sources?}
This dimension primarily distinguishes region-anchored from region-embedded knowledge and has direct implications for multilingual pretraining as low-availability knowledge is less likely to be acquired from English-centric training data alone.
% This dimension is the primary axis distinguishing region-anchored from region-embedded items, and it has direct implications for pretraining: low-availability items are unlikely to be learnable from English-centric data alone, regardless of model scale. 

% \subsection{Utilization of regional knowledge}

\section{The \bench{} Benchmark}

\bench{} is a multilingual benchmark covering 91 languages across 86 countries (127 language--country pairs), spanning 22 scripts and consisting of over 220,000 curated data samples. 
All questions in the dataset are presented in their native languages and scripts, and have a 4-option multiple-choice question (MCQ) answering format. 
Additionally, we provide English translations to facilitate parallel-data analysis. 
\bench{} covers 29 knowledge domains, as summarized in Table \ref{tab:domain_counts}.
The benchmark also covers a wide breadth of linguistic diversity: Ten languages appear across multiple countries, and ten countries contribute data in more than one language (Table \ref{tab:lang_country_diversity} in the Appendix).
This section details the data collection procedure for \bench{}, alongside the subsequent data annotation phase that enables fine-grained model evaluation.
% as well as additional categorical labels we assign to each question in the dataset for the main analysis.

\subsection{Data Collection}
\label{sec:data-collection}

To construct \bench{}, we collect sources of multiple-choice examinations in collaboration with native speakers and regional associations. 
Following a similar protocol to \citet{romanou2024include}, we collect \textit{Academic Exams} spanning diverse subjects (\eg, Humanities \& STEM) and educational levels (\eg, middle \& high school \& university), \textit{Professional Certifications \& Licenses}, and \textit{Regional Licenses} such as exams administered by regional authorities to assess specific qualifications (\eg, driving, marine licenses). 
For high-resource languages where suitable language-specific public benchmarks already exist, we incorporate these resources directly. However, new unpublished data still account for $\sim$78\% of the full benchmark.
During data collection, we design \bench{} to assess multilingual capabilities that extend beyond academic exams to include region-specific knowledge following the taxonomy introduced in Section~\ref{sec:regional_operationalization}.
All collected data undergo native-speaker quality control to verify extraction accuracy, remove context-dependent items that cannot be represented textually, add missing reference context when needed, and assign exam-level metadata. 
Further details on the data collection process and contributor compensation are provided in Appendix~\ref{app:annotation}.

\subsection{Data Annotation}
To enable fine-grained and stratified evaluation, we augment all samples in \bench{} with regional metadata. 
Specifically, we use GPT-5.5 to annotate each question according to the regional knowledge categories introduced in Section~\ref{sec:regional_taxonomy} and the knowledge-availability framework described in Section~\ref{sec:regional_availability}. 
To assess annotation reliability, we conduct a sampled human evaluation in which annotators independently perform the same labeling task. We then measure agreement between human annotations and the LLM-generated labels, and find that agreement rates between humans and LLMs yielded raw exact agreement of 0.760 and an unweighted Cohen's $\kappa$ of 0.676, indicating substantial agreement between humans and LLM raters. Further details of the annotation protocol and agreement analysis are provided in Appendix~\ref{app:annotation}.

We also translate every multiple-choice question, including both the question text and answer options, into English using GPT-5.5. 
These translations allow us to disentangle errors caused by limited access to the underlying knowledge from errors arising from a model's difficulty processing or expressing that knowledge in a particular language. 
% Prior work by \citet{singh2024global} uses parallel multilingual data to study such effects in translated MMLU-style evaluation. 
% In our setting, however, questions are natively sourced across languages and therefore do not share uniform topics, domains, or difficulty levels. \nf{I think no need to compare with global MMLU here.}
We therefore use English translations as a paired reference for each original-language item, enabling controlled within-sample comparisons of model performance across languages.

\begin{table*}[h]
\centering
  \resizebox{\textwidth}{!}{%
\begin{tabular}{l|c>{\columncolor{blue_back}}c|c>{\columncolor{blue_back}}c|c>{\columncolor{blue_back}}c|c>{\columncolor{blue_back}}c|c>{\columncolor{blue_back}}c}
\toprule
\textbf{Model} & \multicolumn{2}{c}{\textbf{Region-agnostic}} & \multicolumn{2}{c}{\textbf{Region-conditional}} & \multicolumn{2}{c}{\textbf{Region-anchored}} &  \multicolumn{2}{c}{\textbf{Region-embedded}} &  \multicolumn{2}{c}{\textbf{All Question Types}} \\

& \multicolumn{1}{c}{In-Language} & \multicolumn{1}{c}{English} 
& \multicolumn{1}{c}{In-Language} & \multicolumn{1}{c}{English} 
& \multicolumn{1}{c}{In-Language} & \multicolumn{1}{c}{English} 
& \multicolumn{1}{c}{In-Language} & \multicolumn{1}{c}{English} 
& \multicolumn{1}{c}{In-Language} & \multicolumn{1}{c}{English} 
              
\\ 
\midrule
               
\multicolumn{10}{c}{\textit{\textit{Commercial Frontier Models}}} \\

Gemini 3.5 Flash & \textbf{81.1} & 75.5 & \textbf{77.7} & 69.7 & \textbf{86.4} & 81.7 & \textbf{75.2} & 63.3 & \textbf{79.7} & 71.2 \\
GPT-5.5  & 78.7 & \textbf{76.0} & 76.2 & \textbf{71.6} & 86.1 & \textbf{82.5} & 72.5 & \textbf{63.4} & 77.5 & \textbf{71.3} \\
GPT-5.4-mini & 72.1 & 73.2 & 64.8 & 64.4 & 79.4 & 80.4 & 62.5 & 55.7 & 70.1 & 66.8 \\
Claude Opus 4.7 & 77.0 & \textbf{76.0} & 75.9 & 67.7 & 83.3 & 82.2 & 67.0 & 60.8 & 74.5 & 70.2 \\

\midrule
\multicolumn{10}{c}{\textit{\textit{Open-Weight Models}}} \\

Apertus-70B & 41.4 & 50.6 & 39.5 & 41.2 & 45.1 & 55.5 & 32.2 & 32.8 & 38.5 & 43.9 \\
Apertus-8B & 39.7 & 48.9 & 36.8 & 41.6 & 43.6 & 54.6 & 32.5 & 33.3 & 38.0 & 42.6 \\

Aya Expanse-32B & 50.0 & 63.0 & 48.8 & 52.0 & 57.2 & 70.0 & 38.7 & 42.2 & 48.0 & 55.1 \\
Aya Expanse-8B & 39.8 & 56.3 & 38.2 & 46.1 & 42.8 & 59.3 & 35.2 & 38.8 & 39.4 & 49.3 \\

Gemma 3-4B & 49.6 & 58.8 & 43.9 & 47.3 & 50.5 & 59.7 & 36.1 & 39.6 & 45.1 & 50.8 \\
Gemma 3-12B & 60.3 & 66.5 & 53.7 & 53.0 & 63.3 & 71.2 & 43.3 & 44.2 & 55.6 & 57.9 \\
Gemma 3-27B & 65.2 & 69.2 & 57.2 & 57.7 & 68.5 & 73.2 & 46.9 & 45.5 & 59.9 & 60.7 \\

Gemma 4-31B & \textbf{70.9} & \textbf{72.1} & \textbf{66.5} & \textbf{65.2} & \textbf{80.0} & \textbf{77.9} & \textbf{57.0} & \textbf{51.0} & \textbf{67.3} & \textbf{64.7} \\

Mistral-7B-instruct & 36.4 & 53.6 & 37.8 & 47.3 & 39.6 & 57.4 & 32.4 & 38.3 & 35.7 & 47.8 \\

Qwen 3-8B & 60.1 & 69.7 & 52.5 & 55.9 & 61.6 & 69.7 & 38.8 & 42.0 & 54.5 & 59.3 \\
Qwen 3-14B & 62.5 & 71.2 & 54.4 & 56.4 & 64.0 & 73.9 & 41.9 & 43.9 & 57.3 & 61.0 \\
Qwen 3-32B & 64.3 & 71.6 & 54.6 & 58.2 & 62.7 & 72.6 & 43.1 & 45.1 & 58.6 & 62.0 \\
Qwen 3.5-27B & 64.0 & 68.8 & 54.3 & 56.1 & 66.9 & 73.3 & 43.0 & 41.6 & 58.3 & 59.1 \\

Tiny Aya Earth & 38.0 & 45.8 & 34.8 & 38.8 & 35.4 & 47.0 & 31.7 & 34.1 & 35.4 & 40.7 \\
Tiny Aya Fire & 39.6 & 49.6 & 40.8 & 40.9 & 38.5 & 49.5 & 32.7 & 35.9 & 37.0 & 43.5 \\
Tiny Aya Global & 36.0 & 46.3 & 33.2 & 40.8 & 35.9 & 46.1 & 30.1 & 33.4 & 34.9 & 40.8 \\
Tiny Aya Water & 39.2 & 48.5 & 36.3 & 42.3 & 36.7 & 47.8 & 31.8 & 34.8 & 36.6 & 42.6 \\

Llama 3.1-70B\_instruct & 56.7 & 66.8 & 55.1 & 58.8 & 65.4 & 73.0 & 45.3 & 45.6 & 54.5 & 59.2 \\
Llama 3.1-8B\_instruct & 37.5 & 53.8 & 36.9 & 43.6 & 38.4 & 57.6 & 29.5 & 35.4 & 34.7 & 47.0 \\

\bottomrule
\end{tabular}
}
  \caption{Average performance (\%) across all languages in \bench{}, both overall and stratified by different regional categories. Quantities are computed as macro-averages across languages. Specific model names/variants can be found in Table \ref{tab:models_evaluated}. In the \textit{In-Language} setting, questions are presented in their native language; in the \textit{English} setting, they are translated into English.  Bold values refer to the best \textit{API} and \textit{open-weight} models per regional category.}
  \label{tab:results}
\end{table*}

% \subsection{\bench{} in a glance}

% We release three versions of the benchmark: \bench{} that contains all the collected data, \bench-base suited for fast multilingual evaluation comprising \textcolor{red}{50000} samples, and \bench-hard, comprising \textcolor{red}{XXXX} hardest samples based on frontier-models' performance (more details in the Appendix \ref{}).
% All variants cover the full set of 91 languages. 
% Additionally, we release the respective \texttt{lm-eval-harness} tasks for universal evaluation, as well as all associated code and experimental results.\looseness=-1

\section{Experimental Setup}
We benchmark \bench{} on commercial models: GPT-5.5, GPT-5.4-mini \citep{singh2025openai}, Claude Opus  4.7,\footnote{\href{https://www.anthropic.com/claude-opus-4-7-system-card}{Claude Opus 4.7 system card}} Gemini-3.5 Flash,\footnote{\href{https://deepmind.google/models/model-cards/gemini-3-5-flash/}{Gemini-3.5 Flash model card}}, and on leading open-weight model families: Llama 3.1 \citep{dubey2024llama}, Qwen 3 \citep{yang2025qwen3}, Qwen 3.5, Gemma 3 \citep{gemma3_2025}, Gemma 4,\footnote{\href{https://deepmind.google/models/gemma/gemma-4/}{Gemma 4 Model}} Mistral \citep{jiang2023mistral7b}, Apertus \citep{apertus2025apertusdemocratizingopencompliant}, Aya Expanse \citep{dang2024ayaexpansecombiningresearch}, and Tiny Aya \citep{salamanca2026tinyayabridgingscale}.
For open-weight and commercial models on multiple-choice benchmarks, we prompt the model to output the answer letter (detailed prompt in the Appendix~\ref{app:prompt}).\footnote{Using log-probabilities to score answer options has been an alternative standard for MCQA evaluations historically \citep{eval-harness}, but as log-probabilities are unavailable for several commercial models in our evaluation, and can introduce systematic biases tied to tokenization and length normalization \citep{Biderman2024LessonsFT}, we decide instead to use the model's ability to generate the correct answer in a specified format as the evaluation target. We acknowledge this choice may conflate the evaluation of regional knowledge recall with fine-grained instruction-following learned during post-training, but the latter is also often a minimal requirement for modern LLMs to be practically useful.}
% \nf{I think the following paragraph should go to the apendix}
% Open-weight model inference was run on an internal cluster of NVIDIA H200 GPUs, with individual runs taking approximately 20 to 60 minutes, for an estimated total of $\sim$160 GPU-hours.
We run Claude Opus 4.7 with greedy decoding and GPT-5 variants with reasoning effort set to low and API-default 
sampling.
% Reproducibility details are provided in Appendix~\ref{app:reproducibility}.

% ===== Add once to preamble (requires \usepackage[most]{tcolorbox}) =====
% \newtcolorbox{rqfinding}{colback=violet!8, colframe=violet!55, arc=3pt,
%   boxrule=0.7pt, left=5pt, right=5pt, top=4pt, bottom=4pt, fontupper=\small}

\section{Results \& Analysis}
We use \bench{} to evaluate current LLMs along four axes: overall performance across languages and regional knowledge types (\S\ref{sec:multilingual-landscape}), whether failures reflect language ability or missing regional knowledge (\S\ref{sec:language_vs_knowledge}), the effect of model scale (\S\ref{sec:scale-comp}), and whether a single aggregate score represents performance at the language and country level (\S\ref{sec:aggregate-vs-local}). As a metric, we report in-language accuracy (i.e., accuracy when a question is presented in its original, native language) macro-averaged across languages unless otherwise noted.

\subsection{Performance summary}
\label{sec:multilingual-landscape}

\begin{rqfinding}
\textbf{Q.} Which models perform best across languages and regional knowledge types?\\[2pt]
\textbf{Finding.} All four frontier models have higher average in-language accuracy than every open-weight model. The best open-weight model, Gemma 4-31B, is 12.4 points below the best frontier model. %Size does not determine the open-weight ranking, as Gemma 3-27B performs better than the larger Qwen 3-32B. 
Every model is less accurate on Region-embedded than on Region-anchored items, by 14.5 points on average for the frontier models and 14.4 for the open-weight models.

%Frontier models have the highest average in-language accuracy, and the open-weight ranking is not determined by size alone. The gap between models' average  in-language accuracy for Region-embedded items vs. Region-anchored items is about 14 points for both frontier and open-weight models.
\end{rqfinding}

\paragraph{Overall performance.}
Frontier models have the highest average in-language accuracy on \bench{} (Table~\ref{tab:results}). The best open-weight model, Gemma 4-31B, is 12.4 points below the best frontier model. Interesting, frontier models tend to have higher in-language accuracy compared to English translation accuracy, while the open-weight models tend to show the reverse phenomenon (with the notable exception of Gemma 4-31B). %Among open-weight models, accuracy does not increase monotonically with parameter count, as Gemma 3-27B has higher in-language accuracy than the larger Qwen 3-32B, and the increase from Qwen 3-14B to the larger Qwen 3-32B is small. Model family and training therefore affect in-language accuracy independently of size.

\paragraph{Performance by regional knowledge type.}
Accuracy is not uniform across regional categories (Table~\ref{tab:results}). For frontier models, in-language accuracy drops by 14.5 points on average from Region-anchored to Region-embedded items. For open-weight models the average drop is 14.4 points. The size of this drop differs by model, with more than 20 points for some open-weight systems (Table~\ref{tab:results}). The drop has the same direction for frontier and open-weight models, which shows that high accuracy on widely documented regional facts does not imply comparable accuracy on knowledge concentrated in local sources. We explore this gap in \S\ref{sec:language_vs_knowledge} by examining whether lower accuracy on Region-embedded items reflects language limitations or gaps in regional knowledge.

%\cm{I think a cleaner conclusion here would be: this suggests something about the training data used for open vs. frontier models. To me, the fact that cultural is often better suggests a larger portion of data with cultural information in training. Probably things like stuff on the web/social media, whereas region-anchored info is maybe more in (pay-walled) news outlets or books. I.e., the stuff that frontier model builders might buy}.
% \jm{frontier models show an average drop of \textbf{14.5} points from region-anchored to region-embedded (in-language), which suggests that region-embedded knowledge is genuinely harder as expected. For open-weight models, we see an average improvement of only \textbf{+1.7 points}, which I think is too small to support strong conclusions; smaller models could also just be guessing. The frontier drop appears to be the more reliable signal here and suggests the difficulty of cultural-specific knowledge.}  \AR{I think I addressed both comments in the section above.}

\subsection{Language versus knowledge}
\label{sec:language_vs_knowledge}

\begin{rqfinding}
\textbf{Q.} When a model answers incorrectly, is the cause weak language ability or missing regional knowledge?\\[2pt]
\textbf{Finding.} Both factors contribute and comparing performance across different prompt languages helps distinguish them. Some regional facts are missed by the models in both the in-language and English setting (termed ``knowledge-dark'' items) and concentrate in specific countries. 
For questions answered correctly in only one setting, frontier and open-weight models differ in which setting they perform better. 
% These findings suggest that limited language proficiency alone cannot explain lower performance on original language questions. 
%Both occur and can be attributed separately. Some regional facts are missed by the model population in both languages (termed ``knowledge-dark'' items) and concentrate in specific countries; some are answered correctly only in English, suggesting weak in-language abilities.
\end{rqfinding}

\begin{figure*}[t]
  \centering
  \includegraphics[width=\textwidth]{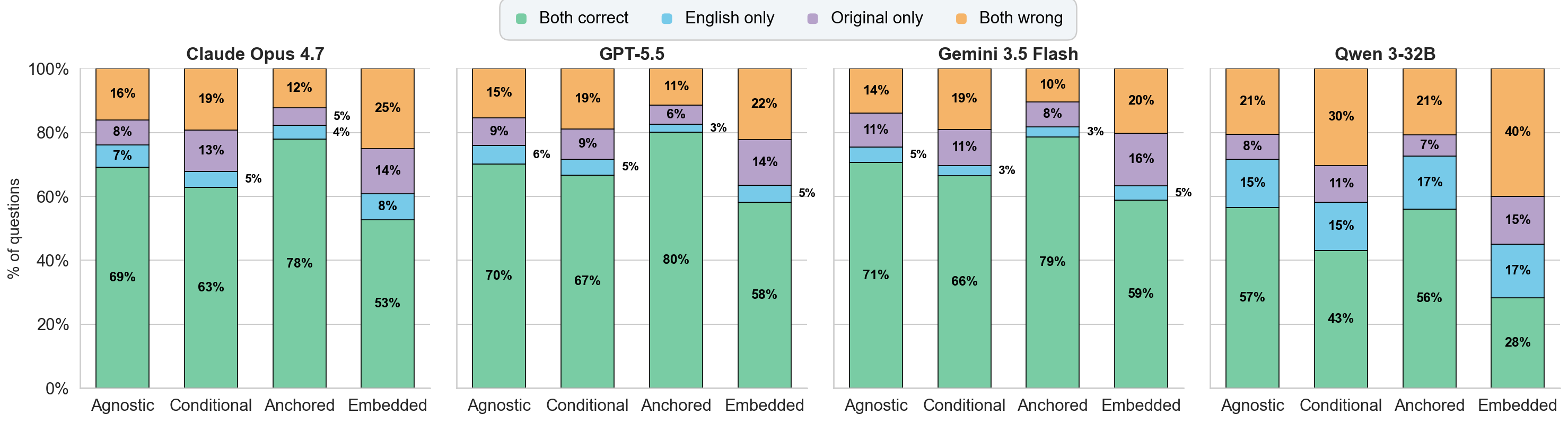}
    \caption{Language-balanced four-outcome decomposition across regional categories.
    For each model, language, and category, item outcomes are partitioned into both correct, English prompt correct only, Original language prompt correct only, and both wrong, then macro-averaged over languages.
    Region-embedded items show the largest shift from both-correct outcomes toward both-wrong outcomes, while the disagreement cells motivate the cross-lingual accessibility analysis.}
  \label{fig:four-outcome}
\end{figure*}

A model can fail to answer a question for many different reasons that are conflated when observing aggregate accuracy alone. For example, an answer may be more indexable from a model's parametric knowledge through a native-language prompt, or it may be accessible in English but lost when processing in the original language due to the model's struggles with basic native language understanding.
To disentangle these effects, we use the parallel English-translated benchmark vs. the original in-language dataset to decompose answer outcomes into four cases: both correct, English-only correct, in-language-only correct, and both wrong.
Figure~\ref{fig:four-outcome} visualizes this decomposition across regional categories, macro-averaged over languages.
The both-wrong cases identify regional facts that appear unretrievable under either language, while the disagreement cases distinguish whether knowledge is more accessible in the original language or only in English, (one of) the model's predominant training languages.

\paragraph{\textit{Knowledge-dark} regional items.}
To identify cases of completely missing regional\footnote{In this study, we constrain our \textit{regional} definition to the Region-anchored and Region-embedded categories, excluding Region-conditional items because failures on them may reflect regional conflation or defaulting to the wrong jurisdiction, rather than lack of access to a fixed regional fact.} knowledge, we define an item as \textit{knowledge-dark} when the 23-model population performs at or below chance in both English and in-language comprehension. Requiring failure both in English and in-language controls for any cross-lingual accessibility issues (a topic we explore next). Items that meet this criterion most plausibly reflect knowledge that is absent from the current model population rather than merely inaccessible.
Knowledge-dark items concentrate in particular countries rather than spreading evenly (Table~\ref{tab:knowledge-dark}).
Across regional-native items the overall dark fraction is 13.6\%, while a few countries exceed 30\%. Among countries with large item counts, the highest fractions are Ghana and Nigeria. While high fractions are also reported for Qatar, Israel, and Hungary, but they are computed on far smaller samples ($n \leq 259$) and thus have large uncertainty. The concentration in specific countries indicates that failures on regional knowledge are localized.

\paragraph{Cross-lingual accessibility.}
We also measure cross-lingual accessibility, where a model answers correctly in only one language but not another.
Such cases suggest that access to the relevant knowledge may depend on the language of the question.
Under this hypothesis, the in-language-only cases---where a model answers correctly in the original language but not in English---correspond to facts that are more easily accessible using the in-language formulation of the question. This pattern is consistent with a model's recall being more successful when the query language matches the language most relevant to the fact, though other factors, such as translation quality, could also produce this pattern. The English-only cases---where the model answers correctly in English but not in the original language---correspond to facts accessible in English but not in the original language, These cases suggest either weaker linguistic abilities or accessibility issues in the original language. Our analyses of such cases are designed to separate these two possible explanations.  Concretely, we summarize the English-only direction with $\Delta=D_{\mathrm{spec}}-D_{\mathrm{agn}}$, corresponding to the English-to-in-language accuracy drop on regional items ($D_{\mathrm{spec}}$) minus the same drop on region-agnostic items ($D_{\mathrm{agn}}$). Subtracting $D_{\mathrm{agn}}$ removes the part of the drop attributable to general difficulty with the original language, so a positive $\Delta$ isolates regional knowledge the model can produce in English but not in the original language. Appendix~\ref{app:access} gives the full definition and the difference-in-differences argument for how $\Delta$ controls for general language ability.

On region-specific items, accuracy is higher in the original language than in English for all four frontier models, by 4.3 to 9.7 points, and higher in English than in the original language for eighteen of the twenty open-weight models, by up to 11.0 points (Table~\ref{tab:cross-lingual}). Among region-specific items answered correctly in exactly one of the two languages, the share correct in the original language only is 61\% to 73\% across the frontier models and 36\% to 61\% across the open-weight models. 
% In short, on the same region-specific items, the language in which a model is more accurate depends on the model group: the four frontier models answer more items correctly in the original language, and 18 of the 20 open-weight models answer more items correctly in English. The direction of the difference is therefore a property of the model, plausibly of its multilingual training data, and not of the items.
Therefore, on the same region-specific items, the direction of the language advantage varies across models, potentially reflecting differences in multilingual training data.
We perform a statistical significance test that supports this result in \cref{app:access}. 

% Difficulty with the original language accounts for most (but not all) of the drop in performance on region-specific items when going from English questions to original language questions. 
To investigate the English-only cases, we compare their prevalence on region-specific and region-agnostic items. 
$D_{\mathrm{spec}}$, the share of region-specific items answered correctly in English but incorrectly in the original language, runs from 9.3\% to 57.5\% across the 24 models. 
After subtracting the corresponding share on region-agnostic items, the remaining difference is at most 5.8 percentage points and is positive for 16 of the 20 open-weight models. 
The similar prevalence of English-only cases on regional and region-agnostic items suggests that general language difficulty plays a substantial role, with a smaller additional gap on regional items.
% After subtracting the accuracy difference on region-agnostic items, at most 5.8 points of the English-to-original-language difference in accuracy remain. This remainder is small but positive for 16 of the 20 open-weight models. 
% Regional knowledge that a model produces only in English is therefore a real but small component of the difference; the larger components are general original-language ability and, for knowledge-dark items, absence of the fact from every model in either language. 

\subsection{Model scale}
\label{sec:scale-comp}

\begin{rqfinding}
\textbf{Q.} Do larger open-weight models show a smaller gap between English and original-language accuracy?\\[2pt]
\textbf{Finding.} Larger open-weight models have higher accuracy on Region-agnostic items, but there is no reliable evidence that the English and in-language gap narrows when scaling up a model.
\end{rqfinding}

In this experiment, we examine whether increased parametric capacity correlates with increased multilingual competence. We restrict the analysis to region-agnostic items, where performance differences are less likely to reflect missing regional knowledge, and to open-weight models (frontier systems are excluded, since they report no public parameter counts).
To perform a regression analysis, we model answer correctness as a function of parameter count,\footnote{We take the log of parameter count and center this variable for numeric stability.} evaluation condition (English vs. in-language), and an interaction term between these two variables.
The interaction term captures whether larger models reduce the English vs. in-language gap (a positive coefficient means the gap narrows as size increases).
We provide details of this formulation in Appendix~\ref{app:analysis_desc} and show the learned coefficients in Table~\ref{tab:rq6_glmm}, where $\beta_1$, $\beta_2$ and $\beta_3$ correspond to parameter count, the evaluation condition, and the interaction term.

Overall, scale improves average accuracy, but we find no reliable evidence that it reduces the English vs.\ in-language gap.
The positive value of the \emph{model size coefficient} $\beta_1$ shows that larger open-weight models have higher accuracy on region-agnostic items, but the \textit{in-language coefficient} $\beta_2$ is negative (with $0$ outside the credible interval\footnote{A credible interval is the Bayesian analogue of a confidence interval; we report credible intervals because the model is fit in a Bayesian framework (ADVI).}), meaning that performance on in-language questions is lower than on English questions at the average model size. The latter result aligns with the results in the previous section about English vs. in-language performance (\S\ref{sec:multilingual-landscape}).  
Finally, the \textit{size $\times$ in-language interaction} term $\beta_3$ is positive but small and uncertain, with $0$ inside the credible intervals for both formulations of the regression model.
The positive point estimate of the $\beta_3$ coefficient is consistent with a slight narrowing of the English--in-language gap as size increases, but its credible interval includes zero, indicating this result is not statistically significant.
% Table~\ref{tab:rq6_family} further shows that scaling behavior differs substantially across model families, with the per-family slope ranging from $0.079$ to $0.567$.
% This suggests the same increase in model size corresponds to very different accuracy gains across families.

\subsection{Aggregate versus local performance}
\label{sec:aggregate-vs-local}

\begin{rqfinding}
\textbf{Q.} Do global model rankings generalize across languages and countries?\\[2pt]
\textbf{Finding.} No. The best model overall (Gemini 3.5 Flash) achieves the highest in-language accuracy in 55 of 91 languages, but is outperformed in the remaining 36, with statistically significant differences in 8.
\end{rqfinding}

\paragraph{Global models may not be the best local models.}
\label{sec:ranking-changes}
Benchmark leaderboards usually report a single global model ranking. We test how well that ranking represents per-language performance by comparing the global leaderboard, ranked by each model's macro-average in-language accuracy, against per-language leaderboards. We summarize the agreement between the global and per-language rankings with the Spearman rank correlation, computed per language and then averaged. Separately, we count the languages whose top-ranked model differs from the global top-ranked model.\footnote{To limit the effect of sampling noise, we resample items within each language and count a language as changed only when its best local model has higher accuracy than the global winner in a significant fraction of resamples.}

Per-language rankings correlate strongly with the global ranking in both settings (Appendix Table~\ref{tab:ranking_stability}), so the relative order of models is largely preserved across languages. The top-ranked model is less stable. In the in-language setting the global winner, Gemini 3.5 Flash, is not the per-language winner for 36 of 91 languages. In the English setting, the global winner, GPT-5.5, changes for 63 of 91 languages. In-language rankings correlate slightly less with the global ranking than English rankings do, indicating more movement below the top rank under in-language evaluation. The five languages with the weakest correlation are Oromo, Tangale, Maithili, Jju, and Ewe (Appendix Table~\ref{tab:rank_stability_worst10}). A global score can therefore name the wrong best model for a given language, which supports reporting per-language results with the global ranking.

\begin{figure}[t]
  \centering
  \includegraphics[width=0.9\columnwidth]{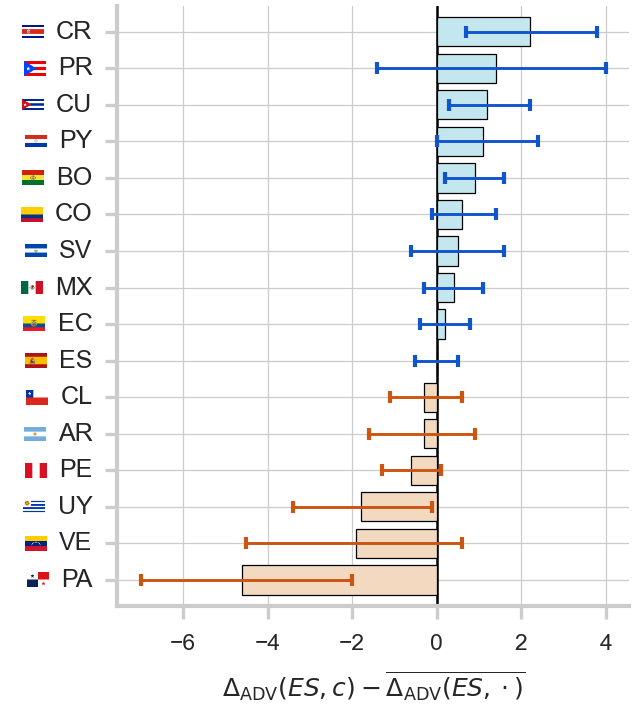}
\caption{Per-region residual of the in-language advantage $\Delta_{\text{ADV}}(\text{ES}, c) - \overline{\Delta_{\text{ADV}}(\text{ES}, \cdot)}$ across Spanish-speaking countries, macro-averaged over all 22 evaluated models with bootstrap 95\% CIs. Positive values indicate that using Spanish-language questions helps more for that country than for all Spanish-speaking regions on average.}  \label{fig:spanish-pluricentric}
\end{figure}

\paragraph{Variation within pluricentric languages.}
\label{sec:pluricentric}
We ask whether the benefit of asking questions in-language is uniform across countries that share the same language.
For each \textit{language--country $(\ell, c)$} pair, we compute the in-language advantage $\Delta_{\mathrm{ADV}} = \mathrm{Acc}_{\mathrm{IL}} - \mathrm{Acc}_{\mathrm{EN}}$, where positive values indicate higher accuracy in the original language than in English.
We then compare each pair's advantage to the average advantage for countries sharing the same language, yielding a gap residual: $\Delta_{\mathrm{ADV}}(\ell,c) - \overline{\Delta_{\mathrm{ADV}}(\ell,\cdot)}$.
A positive gap residual means that in-language querying helps more for that country than for the language on average, while a negative value means it helps less.
We restrict the analysis to region-specific items and estimate uncertainty by bootstrapping items within each \textit{language--country} pair.

Using Spanish as a case study, Figure~\ref{fig:spanish-pluricentric} shows that the in-language advantage varies across Spanish-speaking countries with the language fixed.
For Costa Rica, Cuba, and Bolivia, the in-language advantage is larger than the Spanish average by a margin whose 95\% interval excludes zero. For Panama and Uruguay, it is smaller than the average by such a margin.
The residuals for Puerto Rico, Paraguay, and Venezuela include zero within their 95\% intervals, so they are not distinguishable from the Spanish average.
Because the gap residual is defined from each country's in-language minus English accuracy, it reflects the size of the in-language advantage rather than absolute in-language accuracy.
Appendix Table~\ref{tab:pluricentric-gap-residuals} reports the set of pluricentric languages, including Arabic, Chinese, and Portuguese. Countries with fewer than 30 region-specific items are excluded, which is why only four pluricentric languages appear.
These results indicate that the benefit of querying the model in-language for regional knowledge need not be uniform within a shared language, and that an aggregated language-level score does not capture within-language regional variation.
\section{Related work}

Early Large Language Model (LLM) evaluation relied on English-centric benchmarks such as HellaSwag~\citep{zellers-etal-2019-hellaswag}, MMLU~\citep{hendrycks2021mmlu}, GSM8K~\citep{cobbe2021training}, and BigBench~\citep{srivastava2023beyond}. 
These benchmarks became widely used as tools for evaluating and improving LLM pre-training, but they were all English-only benchmarks, limiting their practical utility for multilingual LLM development.
% While these benchmarks expanded evaluation to complex knowledge comprehension and reasoning, often serving to monitor capabilities during pre-training \citep{penedo2025fineweb2}, their strict reliance on English data limited their utility for multilingual development.

Meanwhile, early multilingual benchmarks were typically restricted to basic linguistic and natural language understanding tasks~\citep{conneau-etal-2018-xnli, ponti-etal-2020-xcopa,ruder-etal-2021-xtreme, liang-etal-2020-xglue, asai2023buffet}. While these datasets enabled multilingual evaluation, they were initially designed to test transfer from high-resource to low-resource languages in finetuning contexts, rather than to understand the complex knowledge and reasoning capabilities of LLMs across languages.
% they primarily focus on transfer from high-resource languages and often rely on translated or parallel data, limiting their ability to capture native, region-specific knowledge.

To bridge the multilingual evaluation gap, recent efforts shifted to translating resources representing complex knowledge from English sources, such MMMLU~\citep{hendrycks2021mmlu}, Global MMLU~\citep{singh2024global}, 
% MMLU-ProX~\citep{xuan2025mmlu}
and PangeaBench~\citep{yue2025pangea}. However, these works evaluated multilingual systems on knowledge from Western ecosystems, as reflected in these originally-English benchmarks.
% Concurrently, a rich body of work has benchmarked LLMs on cultural nuances, societal norms, and localized values using online platforms, synthetic stories, or everyday cross-cultural knowledge
Concurrently, other works filled this gap by benchmarking LLMs on crowdsourced datasets that evaluated the cultural nuances, societal norms, and values linked to local ecosystems~\citep{blend, shi2024culturebank, 10.1145/3715275.3732019, rao2025normad}.
Finally, benchmarks like EXAMS~\citep{hardalov-etal-2020-exams}, \benchvone{}~\citep{romanou2024include}, and Kaleidoscope~\citep{salazar2025kaleidoscope} sourced data from localized academic and professional examinations, explicitly introducing the notion of regionality by evaluating models using resources from specific local environments.
\bench{} expands regional evaluation to 91 languages across 86 countries, providing the broadest coverage yet that enables a detailed analysis of distinct regional knowledge types without conflating geography with culture.

\section{Conclusion}

We introduced \bench{}, a benchmark for evaluating multilingual knowledge across 91 languages, 86 countries and 127 language-country pairs. 
\bench{} comprises native-source questions, sample-level annotations, and English translations to distinguish failures in language ability from gaps in regional knowledge. 
Our results show important variation among categories of regionality, as frontier models struggle most with locally embedded knowledge, and larger open-weight models do not reliably become equally effective across English and original-language prompts. 
These findings motivate evaluation that measures multilingual coverage and also whether models can access and apply knowledge in the contexts where their languages are used.

\section*{Limitations}

% machine-translated approach to get English data
% resouece aboundance for region-anchor and region-embedded is estimated
% difficulty level of questions for a domain across languages is not the same, so comparision between them would be har
While \bench{} substantially expands multilingual and regional evaluation coverage, several limitations remain.

\paragraph{Taxonomy.} First, our distinction between \textit{region-anchored} and \textit{region-embedded} knowledge is operationalized through an annotation heuristic on cross-lingual availability, \ie, whether a fact is reliably documented in globally distributed reference sources, or primarily encoded in native-language and specialized regional materials. 
Consequently, the \textit{region-embedded} category is definitional as items are labeled region-embedded because their supporting evidence is believed by LLM/human annotators to be concentrated outside the English-centric web. 
% The resulting performance gap should therefore be interpreted as a measurement of the magnitude with which models inherit this availability structure, not as evidence that locally grounded topics are intrinsically harder. 
An independent quantitative proxy for availability %(\eg, cross-lingual Wikipedia-edition coverage, multilingual web-frequency estimates, or retrieval-system recall in native-language corpora) 
would let future work calibrate this gap against an external signal rather than the annotation heuristic itself, and we leave this calibration to subsequent work.

\paragraph{Benchmark.} Second, our benchmark adopts a multiple-choice question (MCQ) format exclusively. While MCQs enable scalable evaluation across a large number of languages and domains, they do not fully capture open-ended reasoning, explanation quality, or generative capabilities. Models may sometimes succeed through elimination strategies or pattern matching without demonstrating robust understanding of the underlying regional knowledge. As a result, strong performance on \bench{} should not be interpreted as a complete measure of grounded regional reasoning or multilingual generation ability. We leave the development of open-ended and generative regional knowledge evaluation settings to future work.

Some stages of the benchmark construction pipeline rely on LLM-based processing, including metadata annotation and translation. Although we conduct sampled human validation for the annotation process, systematic biases or inaccuracies introduced by the annotating model may still persist, particularly for low-resource languages or culturally nuanced content. These biases may influence both the categorization of regional knowledge and the downstream analyses derived from those annotations. Future work could reduce these risks through larger-scale expert annotation and multi-annotator validation pipelines across diverse linguistic communities. Finally, all English translations in the benchmark are generated using GPT-5.5 without full human validation, Consequently, translation errors or subtle semantic shifts may still affect these analyses, particularly for culturally nuanced or region-specific concepts.

A discussion of the potential risks of \bench{} can be found in the Appendix \ref{app:risks}.

\section*{Acknowledgments}
We thank Aditi Khandelwal, Deniz Bayazit and Badr Alkhamissi for reading and providing comments on drafts of this paper.
We are grateful to the Masakhane and SomosNLP communities for connecting us with data contributors. We also thank the contributors to the first version of INCLUDE for their foundational work on the benchmark.
We gratefully acknowledge the support of the Swiss National Science Foundation (No. 215390), the AI2050 program at Schmidt Sciences (Grant \#G-25-69783), Sony Group Corporation, and the Swiss National Supercomputing Center (CSCS) in the form of an infrastructure engineering and development project. This research is with support from Google.org and the Google Cloud Research Credits program for the Gemini Academic Program. This project is also funded by the European Union (ERC, RESPECT-LM, 101222478). Views and opinions expressed are however those of the author(s) only and do not necessarily reflect those of the European Union or the European Research Council. Neither the European Union nor the granting authority can be held responsible for them.
This work was supported as part of the Swiss AI Initiative by a grant from the Swiss National Supercomputing Centre (CSCS) under project IDs a139 and a145 on Alps.

\bibliography{custom}

\appendix

\section{Appendix}
\label{sec:appendix}

\subsection{Annotation of \bench{} metadata}
\label{app:annotation}

\subsubsection{Data collection}

All collected data of \bench{} undergo quality control by native speakers. 
After automatic exam extraction, annotators review each question to verify that the question text and answer options were correctly extracted from the source documents.
When extraction errors are identified, annotators manually correct the content using the original materials as reference. 
Annotators also filter out samples that depend on images or tables, and ensure that questions requiring additional context, such as reading-comprehension items, include the relevant reference text in the question field. 
Finally, they assign exam-level metadata to each sample, including the question language, exam topic (in both English and the original language), the academic level when applicable, and country of origin.

The data collection effort targets pluricentric languages and plurilingual countries. Table~\ref{tab:lang_country_diversity} highlights Spanish (20 countries) and Arabic (8) among languages, and India (14 languages) and Nigeria (13) among countries. Full counts are in Table~\ref{tab:lang_country}.

%The benchmark extends INCLUDE, licensed under Apache 2.0. All the new data collected is sourced from publicly available exams and certifications, with licenses verified with the native annotators.

\begin{table}[h]
\centering
\resizebox{\columnwidth}{!}{%
\begin{tabular}{lr||lr}
\toprule
\textbf{Language} & \textbf{\# Countries} & \textbf{Country} & \textbf{\# Languages} \\
\midrule
Spanish    & 20 & India       & 14 \\
Arabic     & 8  & Nigeria     & 13 \\
French     & 2  & Ghana       &  6 \\
Portuguese & 2  & Ethiopia    &  4 \\
Bengali    & 2  & Kenya       &  3 \\
Chinese    & 2  & Mozambique  &  3 \\
Dutch      & 2  & Spain       &  3 \\
Igbo       & 2  & Morocco     &  2 \\
Swahili    & 2  & Pakistan    &  2 \\
Urdu       & 2  & Paraguay    &  2 \\
\bottomrule
\end{tabular}
}
\caption{Languages spanning multiple countries (left) and countries with multiple languages (right).}
\label{tab:lang_country_diversity}
\end{table}

\subsubsection{Human annotator compensation}
\label{app:compensation}

Contributors of newly collected data (\ie, data not already public) choose between two forms of compensation: monetary payment or paper co-authorship. Monetary compensation depends on the number of question--answer (QA) pairs delivered, with bonuses for parsing effort and domain diversity. Table~\ref{tab:annotator-compensation} summarizes the scheme; all amounts are in USD. 

\textbf{Base rate.} We group contributions into four tiers by QA-pair count. The per-pair rate rises from \$0.50 in Tier~1 (300--500 pairs) to \$0.70 in Tier~4 (1{,}201--2{,}000 pairs), so larger contributions earn more per pair.

\textbf{Parsing bonus.} Source exams that require manual parsing before extraction earn an extra \$0.05 per QA pair, added to the tier base rate.

\textbf{Diversity multiplier.} To encourage coverage across knowledge domains (Table~\ref{tab:domain_counts}), we add 5\% to the total payment for contributions spanning up to two domains and 10\% for more than two.

\textbf{Large contributions.} Beyond 2{,}000 QA pairs, the per-pair scheme is replaced by a flat bonus of \$50 for every additional 1{,}000 pairs, added on top of the Tier-4 payment for the first 2{,}000.

For a contribution of $N \leq 2{,}000$ pairs spanning $D$ domains, the total payment is $P = N \cdot (r + b) \cdot m$, where $r$ is the tier base rate, $b{=}\$0.05$ the parsing bonus (0 if not applicable), and $m$ the diversity multiplier ($1.05$ for $D \leq 2$, $1.10$ for $D > 2$). For $N > 2{,}000$, we add \$50 per additional 1{,}000 pairs.

\begin{table}[t]
\centering
\small
\setlength{\tabcolsep}{4pt}
\begin{tabular}{clccc}
\toprule
Tier & \#QA & Base & Parsing & Diversity \\
     &  pairs & (\$/pair) & bonus & multiplier \\
\midrule
1 & 300--500         & 0.50 & 0.05 & +5\,/\,+10\% \\
2 & 501--800         & 0.60 & 0.05 & +5\,/\,+10\% \\
3 & 801--1{,}200     & 0.65 & 0.05 & +5\,/\,+10\% \\
4 & 1{,}201--2{,}000 & 0.70 & 0.05 & +5\,/\,+10\% \\
5 & $>$2{,}000       & \multicolumn{3}{c}{+\$50 per extra 1K} \\
\bottomrule
\end{tabular}
\caption{Compensation scheme for native-speaker contributors, in USD.}
\label{tab:annotator-compensation}
\end{table}

\begin{table}[h]
\centering
\small
\resizebox{\columnwidth}{!}{%
\begin{tabular}{lr}
\toprule
Domain & Count \\
\midrule
Language and Linguistics & 28,000 \\
History & 20,936 \\
Biology & 17,067 \\
Law & 15,027 \\
Mathematics & 14,065 \\
Medical Sciences & 13,326 \\
Environmental and Earth Sciences & 11,952 \\
Physics & 10,062 \\
Literature & 9,953 \\
Chemistry & 7,731 \\
Political Science & 7,547 \\
Religious Studies & 7,418 \\
Economics & 7,327 \\
Vocational and Licensing Exams & 7,081 \\
Business and Management & 6,546 \\
Anthropology & 5,732 \\
Philosophy & 5,076 \\
General Knowledge & 4,797 \\
Computer Science & 4,363 \\
Sociology & 4,144 \\
Psychology & 3,200 \\
Education & 2,921 \\
Engineering & 2,379 \\
Performing Arts & 982 \\
Visual Arts & 921 \\
Allied Health & 538 \\
Technical and Trade Education & 405 \\
Geography & 133 \\
Journalism, Media Studies, and Communication & 47 \\
\bottomrule
\end{tabular}%
}
\caption{Unique domains and their distribution in \bench{}.}
\label{tab:domain_counts}
\end{table}

\subsubsection{LLM annotation}
For each question in \bench{}, we use GPT-5.5 for two annotation steps. 
First, we translate the original-language question and answer options into English using a structured prompt designed for multiple-choice assessment items (Figure~\ref{fig:app-translation-prompt}). 
The prompt instructs the model to preserve the item's meaning, difficulty, answer-choice order, labels, formatting, and any explicit answer markers, and to return a constrained JSON output.

Second, we classify each item into four regional-knowledge categories: region-agnostic, region-conditional, region-anchored, or region-embedded. 
We use a separate annotation prompt that defines the categories, specifies an ordered decision procedure, and asks the model to classify the cognitive target of the item rather than relying on surface cues such as the source language or country (Figure~\ref{fig:app-regional-annotation-prompt}). 
This annotation pipeline provides both an English version of each item for cross-lingual evaluation and metadata describing the type of regional knowledge required to answer it.

\subsubsection{Reliability of the LLM regional annotations}
\label{app:annotation-reliability}

To validate the regional labels, we run a sampled human study in which two AI expert annotators independently repeat the labeling task, and we measure agreement between the annotators and between the annotators and the LLM.

\paragraph{Sampling protocol.} We draw a stratified sample of 25 items per category (100 in total), allowing at most one item per language within each category to spread coverage across the benchmark. Annotators see the English translation, so the task is language-independent. Items are shuffled and interleaved across categories, and each annotator works from a separate sheet with the LLM label hidden, keeping the two annotations independent and blind. Both annotators receive the same category definitions and item information (question, options, and country) given to the LLM.

\paragraph{Inter-annotator agreement.} The four categories are nominal, so we report nominal-label agreement metrics: raw exact agreement, unweighted Cohen's $\kappa$, and Krippendorff's $\alpha$ with a nominal distance function. Agreement between the two annotators is high, with a raw exact agreement of 0.760, an unweighted Cohen's $\kappa$ of 0.676, and a Krippendorff's $\alpha$ of 0.677. These values indicate that the category definitions are applied consistently.

\paragraph{Human--LLM agreement.} We then compare the LLM labels with the human annotations. Averaged over the two annotators, the LLM matches the human label with a raw exact agreement of 0.795 and an unweighted Cohen's $\kappa$ of 0.727. Treating the two annotators and the LLM as three raters, Krippendorff's $\alpha$ is 0.710. On the 76 items where both annotators agree, the LLM matches their consensus label 0.895 of the time. Overall, the LLM labels align with human judgments, supporting their use as sample-level annotations in our analyses.

%%%%%%%%%%%%%%%%%%%%%%%%%%%%%%%%%%%%%%%%%%%%%%%%%%%%%%%%%%%%%%%

\begin{table}[t]
\centering
\small
\resizebox{\columnwidth}{!}{%
\begin{tabular}{lr}
\toprule
\textbf{Model ID} & \textbf{Size (B)} \\
\midrule
CohereLabs/tiny-aya-earth          & 3.35 \\
CohereLabs/tiny-aya-fire           & 3.35 \\
CohereLabs/tiny-aya-global         & 3.35 \\
CohereLabs/tiny-aya-water          & 3.35 \\
CohereLabs/aya-expanse-8b          & 8.03 \\
CohereLabs/aya-expanse-32b         & 32.30 \\
google/gemma-7b-it                 & 8.54 \\
google/gemma-3-4b-it               & 4.30 \\
google/gemma-3-12b-it              & 12.00 \\
google/gemma-3-27b-it              & 27.43 \\
google/gemma-4-31B-it              & 32.68 \\
meta-llama/Llama-3.1-8B-Instruct   & 8.03 \\
meta-llama/Llama-3.1-70B-Instruct  & 70.55 \\
mistralai/Mistral-7B-Instruct-v0.3 & 7.25 \\
Qwen/Qwen3-8B                      & 8.19 \\
Qwen/Qwen3-14B                     & 14.77 \\
Qwen/Qwen3-32B                     & 32.76 \\
Qwen/Qwen3.5-27B                   & 27.78 \\
swiss-ai/Apertus-8B-Instruct-2509  & 8.05 \\
swiss-ai/Apertus-70B-Instruct-2509 & 70.60 \\
\bottomrule
\end{tabular}
}
\caption{Models evaluated on the \bench{} benchmark, sorted by model family and parameter count.}
\label{tab:models_evaluated}
\end{table}

%%%%%%%%%%%%%%%%%%%%%%%%%%%%%%%%%%%%%%%%%%%%%%%%%%%%%%%%%%%%%%%

\begin{table*}[h]
\centering
\resizebox{1\textwidth}{!}{%
\begin{tabular}{llr|llr|llr}
\toprule
\textbf{Language} & \textbf{Country} & \textbf{N} & \textbf{Language} & \textbf{Country} & \textbf{N} & \textbf{Language} & \textbf{Country} & \textbf{N} \\
\midrule
Albanian & Albania & \textbf{2,317} & Georgian & Georgia & \textbf{546} & Sena & Mozambique & \textbf{488} \\
Amharic & Ethiopia & \textbf{1,864} & German & Germany & \textbf{1,912} & Serbian & Serbia & \textbf{1,412} \\
Arabic &  & \textbf{9,459} & Greek &  & \textbf{2,279} & Sindhi & Pakistan & \textbf{314} \\
 & Egypt & 1,616 &  & Cyprus & 42 & Sinhala & Sri Lanka & \textbf{290} \\
 & Jordan & 4,749 &  & Greece & 2,237 & Slovak & Slovakia & \textbf{130} \\
 & KSA & 71 & Guaraní & Paraguay & \textbf{2,897} & Somali & Ethiopia & \textbf{264} \\
 & Kuwait & 18 & Gujarati & India & \textbf{321} & Spanish &  & \textbf{14,672} \\
 & Morocco & 193 & Hausa & Nigeria & \textbf{1,490} &  & Argentina & 887 \\
 & Palestine & 623 & Hebrew & Israel & \textbf{2,282} &  & Belice & 15 \\
 & Qatar & 547 & Hindi & India & \textbf{4,804} &  & Bolivia & 913 \\
 & UAE & 107 & Hungarian & Hungary & \textbf{1,386} &  & Chile & 689 \\
Armenian & Armenia & \textbf{841} & Ibibio & Nigeria & \textbf{272} &  & Colombia & 1,342 \\
Assamese & India & \textbf{272} & Idoma & Nigeria & \textbf{642} &  & Costa Rica & 347 \\
Azerbaijani & Azerbaijan & \textbf{6,143} & Igala & Nigeria & \textbf{303} &  & Cuba & 584 \\
Basque & Spain & \textbf{588} & Igbo &  & \textbf{1,475} &  & Ecuador & 1,807 \\
Belarusian & Belarus & \textbf{489} &  & Nigeria & 1,434 &  & El Salvador & 453 \\
Bengali &  & \textbf{14,694} &  & Nigeria & 41 &  & España & 4,327 \\
 & Bangladesh & 14,388 & Indonesian & Indonesia & \textbf{11,045} &  & Jamaica & 19 \\
 & India & 306 & Italian & Italy & \textbf{4,608} &  & México & 1,090 \\
Bulgarian & Bulgaria & \textbf{2,827} & Japanese & Japan & \textbf{2,541} &  & Nicaragua & 20 \\
Chinese &  & \textbf{12,425} & Jju & Nigeria & \textbf{599} &  & Panamá & 48 \\
 & China & 12,005 & Kannada & India & \textbf{541} &  & Paraguay & 214 \\
 & Hong Kong & 405 & Kazakh & Kazakhstan & \textbf{5,591} &  & Perú & 1,213 \\
 & Singapore & 15 & Kinyarwanda & Rwanda & \textbf{381} &  & Puerto Rico & 251 \\
Croatian & Croatia & \textbf{2,369} & Korean & Korea & \textbf{1,607} &  & República Dominicana & 21 \\
Czech & Czechia & \textbf{714} & Lithuanian & Lithuania & \textbf{682} &  & Uruguay & 263 \\
Dagbani & Ghana & \textbf{1,160} & Luo & Kenya & \textbf{2,194} &  & Venezuela & 169 \\
Dangme & Ghana & \textbf{569} & Macedonian & North Macedonia & \textbf{1,994} & Swahili &  & \textbf{491} \\
Danish & Denmark & \textbf{446} & Maithili & India & \textbf{245} &  & Kenya & 482 \\
Darija & Morocco & \textbf{2,956} & Makhuwa & Mozambique & \textbf{254} &  & Tanzania & 9 \\
Dogri & India & \textbf{263} & Malay &  & \textbf{998} & Swedish & Sweden & \textbf{2,494} \\
Dutch &  & \textbf{1,446} &  & Malaysia & 986 & Tagalog & Philippines & \textbf{492} \\
 & Belgium & 521 &  & Singapore & 12 & Tamil & India & \textbf{1,215} \\
 & Netherlands & 925 & Malayalam & India & \textbf{886} & Tangale & Nigeria & \textbf{273} \\
Ekpeye & Nigeria & \textbf{300} & Marathi & India & \textbf{304} & Telugu & India & \textbf{11,237} \\
Embu & Kenya & \textbf{294} & Nepali & Nepal & \textbf{1,316} & Tigrinya & Ethiopia & \textbf{1,901} \\
Esan & Nigeria & \textbf{152} & Nyanja & Mozambique & \textbf{997} & Turkish & Turkey & \textbf{2,765} \\
Estonian & Estonia & \textbf{490} & Obolo & Nigeria & \textbf{1,663} & Twi & Ghana & \textbf{1,568} \\
Ewe & Ghana & \textbf{350} & Oriya & India & \textbf{372} & Tyap & Nigeria & \textbf{1,395} \\
Fante & Ghana & \textbf{1,068} & Oromo & Ethiopia & \textbf{993} & Ukrainian & Ukraine & \textbf{1,421} \\
Finnish & Finland & \textbf{838} & Persian & Iran & \textbf{24,102} & Urdu &  & \textbf{623} \\
French &  & \textbf{1,785} & Polish & Poland & \textbf{1,928} &  & India & 111 \\
 & Canada & 846 & Portuguese &  & \textbf{1,294} &  & Pakistan & 512 \\
 & France & 939 &  & Brazil & 425 & Uzbek & Uzbekistan & \textbf{2,270} \\
Fula & Nigeria & \textbf{261} &  & Portugal & 869 & Vietnamese & Vietnam & \textbf{8,321} \\
Ga & Ghana & \textbf{1,837} & Punjabi & India & \textbf{742} & Yoruba & Nigeria & \textbf{1,346} \\
Galician & España & \textbf{350} & Russian & Russia & \textbf{6,241} &  &  &  \\
\bottomrule
\end{tabular}%
}
\caption{Language and country distribution of \bench{}.}
\label{tab:lang_country}
\end{table*}

\subsection{Inference details}
\label{app:inference}

%- model names \\
%- inference prompt \\
%- inference settings \\

\subsubsection{Models}
\label{app:models}

Table~\ref{tab:models_evaluated} lists the 19 instruction-tuned checkpoints evaluated, covering 4B to 70B dense parameters. All checkpoints are public on the Hugging Face Hub and are loaded by their repository ID at the current revisions as of 2026-05-23.

\subsubsection{Prompt format}
\label{app:prompt}

All models are evaluated zero-shot in a chat-template setting using the following system instruction:

\begin{tcolorbox}[
  colback=gray!5,
  colframe=black,
  title={Inference prompt for multiple-choice \\answer extraction},
  boxrule=0.5pt,
  arc=2mm,
  outer arc=2mm,
  boxsep=1mm,
  left=1mm,
  right=1mm,
  top=1mm,
  bottom=1mm,
  fontupper=\footnotesize,
  breakable
]
\label{app-inference-prompt}

\vspace{0.4em}

{\ttfamily
\begin{minipage}{\linewidth}
You will be given a multiple-choice question from \{country\}. Before responding, analyze the question and answer options. You can do any reasoning; however, you MUST put your answer inside the answer tag: \textless answer\textgreater A/B/C/D\textless/answer\textgreater. Do not include any other text on that line.
\end{minipage}
}

\end{tcolorbox}

The task instruction contains the question followed by the four options labelled \texttt{A.}-\texttt{D.} and a trailing \texttt{Answer:} cue.

We apply each model's chat template via \texttt{lm\_eval}'s \texttt{-{}-apply\_chat\_template} flag. We also set \texttt{-{}-enable-thinking=True} for the model families that support it, namely Apertus-70B, Qwen 3, Qwen 3.5, and Gemma 4.

Alternativly, using log-probabilities to score answer options has been an  standard for MCQA evaluations historically \citep{eval-harness}, but as log-probabilities are unavailable for several commercial models in our evaluation, and can introduce systematic biases tied to tokenization and length normalization \citep{Biderman2024LessonsFT}, we decide instead to use the model's ability to generate the correct answer in a specified format as the evaluation target. We acknowledge this choice may conflate the evaluation of regional knowledge recall with fine-grained instruction-following learned during post-training, but the latter is a minimal requirement for modern LLMs.

\subsubsection{Answer extraction and scoring}
Each generation is post-processed by two custom \texttt{lm\_eval} filters and graded by the harness's \texttt{exact\_match} metric (case- and punctuation-insensitive). We initially scored with strict extraction (\texttt{em\_hard}) but, prompted by the response-pattern analysis below, softened the parsing into a fallback scorer (\texttt{em\_soft}). We calculate both:

\paragraph{em\_hard} takes the letter inside the first \texttt{<answer>$X$</answer>} tag with $X \in \{A, B, C, D\}$ (case insensitive). Responses without a well-formed tag score zero.

%\textbf{em\_soft} adds a fallback applied in order: (i)~\emph{Tag: Letter}, the strict tag above (\texttt{<answer>A</answer>}); (ii)~\emph{Tag: Letter + Text}, the first \texttt{A}--\texttt{D} letter inside the tag (\texttt{<answer>A.</answer>}); (iii)~\emph{No Tag: Letter}, a bare letter (\texttt{"A"}); (iv)~\emph{No Tag: Letter + Punct. + Text}, a leading letter followed by punctuation and text (\texttt{"A. Madrid"}); (v)~\emph{No Tag: The Answer is X}, explicit phrasing (\texttt{"the answer is A"}). Responses caught by no stage fall into \emph{Other} or \emph{Empty} and score zero.

\begin{figure*}[t!]
    \centering
    \includegraphics[width=1\linewidth]{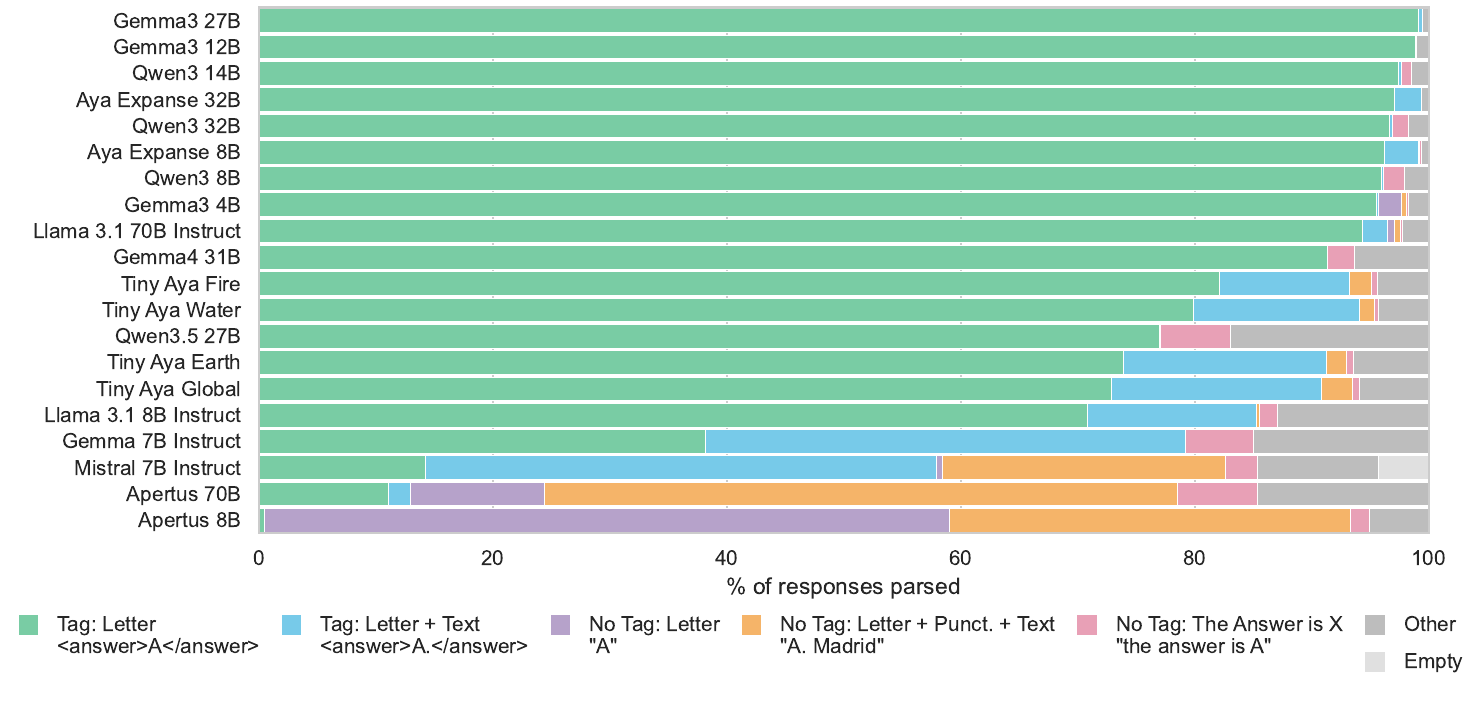}
    \caption{Response pattern distribution per model: share of responses caught at each stage of the \emph{em\_soft} fallback pipeline. Models are ordered by their strict tag share. Counts pool local-language and English questions.}
    \label{fig:response-patterns}
\end{figure*}

\paragraph{em\_soft} adds a fallback applied in order:
\begin{enumerate}
    \item \emph{Tag:~Letter},~strict~tag~(\texttt{<answer>$A$</answer>}).
    \item \emph{Tag: Letter + Text}, the first \texttt{A}--\texttt{D} letter inside the tag (\texttt{<answer>$A$.</answer>}).
    \item \emph{No Tag: Letter}, a bare letter (\texttt{``$A$''}).
    \item \emph{No Tag: Letter + Punct. + Text}, a leading letter followed by punctuation and text.
    \item \emph{No Tag: Answer is $X$}, explicit phrasing.
    \item Responses caught by no stage fall into \emph{Other} or \emph{Empty} and score zero.
\end{enumerate}

Appendix~Figure~\ref{fig:response-patterns} shows the share of responses caught at each stage per model.

\subsubsection{Hardware and Environment}
\label{app:env}

All jobs are executed on the Alps research infrastructure at CSCS on Grace-Hopper (GH200) nodes with 4 GPUs and 460\,GB of host RAM per node. The evaluation of all the reported results sums up to 600 GPU hours. Each \texttt{(model, variant)} evaluation runs as a SLURM job using single-node tensor parallelism (\texttt{tensor\_parallel\_size=4}). For the 27--70B models the 128-subtask group is sharded across 2 or 4 independent jobs to reduce wall-time; per-task results are concatenated post-hoc. The full sweep uses \texttt{vllm 0.20.0}, \texttt{transformers 5.8.1}, and \texttt{lm\_eval 0.4.12}.

\subsubsection{Reproducibility}
\label{app:reproducibility}

The full evaluation pipeline, including task definitions, prompts, answer extraction filters, and model configurations, will be
released on GitHub. The per-sample model results will be released on Hugging Face.

A single \texttt{(model, variant)} run
can be reproduced with the following command:

%\begin{figure}[h!]
\begin{verbatim}
lm_eval \
  --model vllm \
  --model_args "pretrained=<HF_MODEL_ID>,
      dtype=auto,
      gpu_memory_utilization=0.75,
      tensor_parallel_size=4" \
  --tasks include_cot_reasoning_en \
  --apply_chat_template \
  --system_instruction 
      "<SYSTEM_INSTRUCTION>" \
  --gen_kwargs max_gen_toks=4096 \
  --output_path
      results/include/<MODEL>/en \
  --log_samples \
  --batch_size auto
\end{verbatim}
%\end{figure}

% technically
%   --include_path include-tasks/include_cot_reasoning \

Replacing \texttt{include\_cot\_reasoning\_en} with
\texttt{include\_cot\_reasoning\_og} runs the original-language
variant.

\clearpage
\begin{figure*}[p]
\centering
\begin{tcolorbox}[
  enhanced jigsaw,
  colback=gray!5,
  colframe=black,
  title={Translation prompt},
  boxrule=0.5pt,
  arc=2mm,
  outer arc=2mm,
  boxsep=1.5mm,
  left=2mm,
  right=2mm,
  top=1mm,
  bottom=1mm,
  width=0.96\textwidth,
  fontupper=\footnotesize
]
% \label{fig:app-translation-prompt}

\textbf{\underline{SYSTEM INSTRUCTION}}

\vspace{0.4em}

{\ttfamily\footnotesize
\begin{minipage}{0.98\linewidth}
You are a professional translator specializing in assessment items (MCQs).\\
Translate into natural, publication-quality English while preserving difficulty,\\
logic, and fairness. Do not explain your choices.

\vspace{0.8em}

\textbf{\underline{USER PROMPT}}

\vspace{0.4em}

Source language: \{\}

\vspace{0.6em}

Task: Translate the following multiple-choice question and its options into English.

\vspace{0.6em}

Rules (strict):\\
1) Preserve meaning, tone, register, and difficulty. Do NOT add hints or explanations.\\
2) Do NOT add, remove, merge, split, or reorder choices.\\
3) Preserve any existing labels exactly (\eg, ``A.'', ``B)'', ``\textcircled{\scriptsize 1}'', ``a)'').\\
\hspace*{1.5em}If choices have no labels, do NOT introduce labels.\\
4) Keep all names, numbers, formulas, units, URLs, emojis, hashtags, @mentions as-is unless a standard English convention clearly applies\\
\hspace*{1.5em}(\eg., decimal comma $\rightarrow$ decimal point, date format to English).\\
5) Preserve structure and inline formatting: line breaks, Markdown/HTML/LaTeX, code blocks, variables like \{\{...\}\}, and tags like \textless...\textgreater.\\
\hspace*{1.5em}Do NOT translate content inside code blocks, variables, or tags.\\
6) Translate idioms and fixed expressions idiomatically (not word-for-word).\\
7) If a phrase is genuinely ambiguous in a way that could affect the answer, choose the best rendering and append the alternative in square brackets [like this] immediately after the phrase.\\
8) Do NOT infer or reveal the correct answer. If the source includes an explicit correct-answer marker, keep it exactly as written.\\
9) Output must be valid JSON with exactly these keys and nothing else:

\vspace{0.4em}

\hspace*{1.5em}\{\\
\hspace*{3.0em}``question\_en'': ``\textless translated question text\textgreater'',\\
\hspace*{3.0em}``choices\_en'': [``\textless choice 1\textgreater'', ``\textless choice 2\textgreater'', ...]\\
\hspace*{1.5em}\}

\vspace{0.4em}

\hspace*{1.5em}Do not include any extra fields unless ambiguity required rule 7; if so, add\\
\hspace*{1.5em}``notes'': ``\textless one concise sentence about the ambiguity\textgreater''.

\vspace{0.8em}

Input JSON:\\
\hspace*{1.5em}\{\\
\hspace*{3.0em}``question'': ``\{\}'',\\
\hspace*{3.0em}``choices'': [\\
\hspace*{4.5em}``\{\}'',\\
\hspace*{4.5em}``\{\}'',\\
\hspace*{4.5em}``\{\}'',\\
\hspace*{4.5em}``\{\}''\\
\hspace*{3.0em}]\\
\hspace*{1.5em}\}

\vspace{0.6em}

Output JSON only, no preface, no prose.
\end{minipage}
}

\end{tcolorbox}
\caption{Translation prompt used to create the English version of each multiple-choice item.}
\label{fig:app-translation-prompt}
\end{figure*}
\clearpage

\clearpage
\begin{figure*}[p]
\centering
\begin{tcolorbox}[
  enhanced jigsaw,
  colback=gray!5,
  colframe=black,
  title={Regional-knowledge annotation prompt},
  boxrule=0.5pt,
  arc=2mm,
  outer arc=2mm,
  boxsep=1.5mm,
  left=2mm,
  right=2mm,
  top=1mm,
  bottom=1mm,
  width=0.96\textwidth,
  fontupper=\scriptsize
]

\textbf{\underline{SYSTEM INSTRUCTION}}

\vspace{0.4em}

{\ttfamily\scriptsize
\begin{minipage}{0.98\linewidth}
You are an expert annotator for a multilingual evaluation benchmark. Your job is to classify exam-style multiple-choice questions by the kind of regional or cultural knowledge they test.

Be precise and consistent. Follow the decision procedure exactly and apply the first matching category.

Classify the question based on its cognitive target---what knowledge or reasoning is required to answer it---not merely on surface nouns, the source region, or the language of the item. The same topic can belong to different categories depending on what the question asks.

Use the provided correct answer only to understand the intended question, not to override the classification rules.

\vspace{0.9em}

\textbf{\underline{USER PROMPT}}

\vspace{0.4em}

\# Task\\
Classify the item into exactly one category: A, B, C, or D.

\vspace{0.7em}

\# Categories

\vspace{0.4em}

\textbf{\#\# A --- Region-agnostic}\\
The correct answer does not depend on region, country, jurisdiction, or local cultural context.

Typical cases: mathematics; physics and chemistry laws; formal logic; anatomy and physiology; mechanism-of-action pharmacology; abstract reasoning; internationally standardized scientific facts.

Key test: Would the answer remain the same regardless of where the question is asked? If yes, and the item is not specifically about a region, classify as A.

\vspace{0.7em}

\textbf{\#\# B --- Region-conditional}\\
The correct answer depends on the region, country, jurisdiction, or institutional system the question is anchored to.

Typical cases: laws and regulations; legal ages; tax thresholds; prescription rules; voting eligibility; professional licensing; school grading systems; civil procedure; speed limits; statutes of limitation.

Key test: Could the same question template plausibly be asked in multiple regions and yield different correct answers because local rules differ? If yes, classify as B.

Important: B takes priority over all other categories. A question about a specific country's law is still B if the answer is jurisdiction-dependent in principle.

\vspace{0.7em}

\textbf{\#\# C --- Region-anchored factual}\\
The question is about a specific region, country, people, or place, and the answer is unique and fixed rather than jurisdiction-variable. The relevant knowledge is typically available in globally distributed or widely accessible reference sources.

Typical cases: capitals; heads of state; major national historical events; major wars or treaties; large-scale demographics; internationally known figures; widely documented landmarks or institutions.

Key test: Is the topic region-specific, with one fixed answer that is typically findable in English-language or internationally distributed sources? If yes, classify as C.

\vspace{0.7em}

\textbf{\#\# D --- Region-embedded cultural/practice knowledge}\\
The question has a unique and fixed answer, but the knowledge is deeply embedded in a particular language community, regional practice, or local cultural tradition, and is typically sparse in globally distributed sources.

Typical cases: traditional music, rhythms, instruments, or dances; regional cuisine specifics and culinary techniques; folk customs and vernacular practices; local mythology or ritual traditions; regionally specialized terminology; local historical figures of limited international prominence; knowledge that generally requires native-language, community-specific, or specialized regional sources.

Key test: Would answering this usually require native-language materials, local cultural familiarity, or specialized regional sources rather than broadly accessible global references? If yes, classify as D.

\vspace{0.7em}

\# Decision procedure\\
Apply these steps in order. Stop at the first match.

1. Does the correct answer depend on which region, jurisdiction, or local institutional system the question is anchored to? If yes, classify as B.\\
2. If not, is the question region-independent and about universal scientific, mathematical, formal, or abstract knowledge? If yes, classify as A.\\
3. If not, is the question about a specific region, country, place, people, or tradition, with a unique fixed answer that is typically available in globally distributed reference sources? If yes, classify as C.\\
4. Otherwise, classify as D.

\vspace{0.7em}

\# Output format\\
Return valid JSON only. Do not include markdown fences, preamble, explanations outside JSON, or trailing prose.

Use exactly this schema:

\vspace{0.3em}

\hspace*{1.5em}\{\\
\hspace*{3.0em}``reasoning'': ``\textless 2--3 concise sentences applying the decision procedure\textgreater'',\\
\hspace*{3.0em}``category'': ``\textless A | B | C | D\textgreater'',\\
\hspace*{3.0em}``confidence'': ``\textless high | medium | low\textgreater'',\\
\hspace*{3.0em}``notes'': ``\textless edge-case considerations, plausible alternative category, or translation concerns; use an empty string if none\textgreater''\\
\hspace*{1.5em}\}

\vspace{0.7em}

\# Item to annotate\\
Source region: \{source\_region\}\\
Source language: \{source\_language\}\\
Question (native): \{question\_native\}\\
Question (English): \{question\_english\}\\
Options (native): \{options\_native\}\\
Options (English): \{options\_english\}
\end{minipage}
}

\end{tcolorbox}
\caption{Prompt used to classify each item into regional-knowledge categories and assign the pluricentric flag.}
\label{fig:app-regional-annotation-prompt}
\end{figure*}
\clearpage

\section{Data Analysis Setup}\label{app:analysis_desc}
\subsection{Multilingual Knowledge Accessibility}\label{app:access}
Intuitively, a raw drop in accuracy on regional items can confound two causes we want to separate: generally weaker processing in that language, or inaccessiblility of knowledge in that language. The region-agnostic items let us quantify the first case; they require the same local-language processing as for the region-specific questions, but contain no regional content, so absence of regional knowledge is not be a factor for these cases. We can therefore use this in-language--English difference in accuracy as a control. Explicitly, we use  a difference-in-differences approach rather than computing raw differences: we compare the drop in in-language--English accuracy on regional items to the same drop on region-agnostic items. 
$\Delta = D_{\text{spec}} - D_{\text{agn}}$ is the part of the accuracy drop we thus attribute to inaccessibility  of regional knowledge.

\paragraph{Significance of the in-language versus English asymmetry.}
Section~\ref{sec:language_vs_knowledge} reports that the frontier models are more accurate in the original language on region-specific items and the open-weight models more accurate in English. Because per-language item counts are small, we test that asymmetry separately within each combination of one model and one language rather than only on the pooled data. For a given model and language we count the region-specific items answered correctly in the original language and incorrectly in English, and the items answered correctly in English and incorrectly in the original language; items answered correctly in both languages or in neither enter neither count and are excluded. Under the null hypothesis, an item answered correctly in exactly one of the two languages is equally likely to fall into either count, which gives an exact two-sided McNemar test. We apply it at $\alpha=0.05$ to every combination holding at least one such item and report the two rejection directions separately in Table~\ref{tab:cross-lingual} (columns \emph{IL larger}, \emph{EN larger} and \emph{Tested}).

For the four frontier models, 355 combinations can be tested and 210 reject the null. The original-language count is the larger one in 208 of those 210, spread over between 44 and 60 of the 88 or 89 languages tested for each model. For the twenty open-weight models, 1{,}800 combinations can be tested and 855 reject the null; the English count is the larger one in 674 of those 855. The test describes which of the two counts is larger for a given model and language. It does not attribute that asymmetry to a particular cause, and translation quality of the English variant, pretraining mixture, prompt template sensitivity and tokenisation are all consistent with the same observation.

\subsection{Scaling vs. Multilingual Competence}
We provide further details on the regression models used in the analysis in Section~\ref{sec:scale-comp}. 
All regression models include a per-family random intercept and slope on parameter count to reflect that accuracy for a given family may scale at its own rate, and a per-language random intercept to capture the model's. In our primary regression model, we also include an item-level random intercept.\footnote{ An aggregated binomial model over all region-agnostic items, dropping the item random intercept, yields very similar coefficients.} In words, the random effects account for the data being grouped: items belong to languages and models to families, so outcomes within a group are correlated rather than independent. A separate intercept for each family, language, and item captures the baseline differences in accuracy between groups, and the family-level slope allows the effect of size to differ from one family to the next. Without these terms, we would be breaking the independence assumptions made by regression models; the model would treat correlated observations as independent and report confidence intervals that are too narrow.
\newcommand{\size}[1]{P_{#1}}              % parameter count
\newcommand{\lsize}[1]{\log P_{#1}}        % centered log parameter count
\newcommand{\cond}{\mathrm{cond}}
\newcommand{\famof}[1]{f(#1)}
\newcommand{\langof}[1]{\ell(#1)}
\newcommand{\bfam}[2]{b^{\mathrm{fam}}_{#1,#2}}  
Putting this together, we get the following equation:
\begin{equation}
\begin{aligned}
\logit \pi_{m,i,c}
={}& \beta_0 + b_{0,\famof{m}} + u_{\langof{i}} + v_i \\
&+ \bigl(\beta_1 + b_{1,\famof{m}}\bigr) \lsize{m} \\
&+ \beta_2\,c + \beta_3\,c\lsize{m} .
\end{aligned}
\end{equation}
or using \texttt{lmer} notation:
\begin{align*}
    &\texttt{correct \textasciitilde{} logP * c + (1 + logP \!\!|\!\! family)}\\
       &\quad\quad\quad\quad\texttt{ + (1 | language) + (1 | item)}
\end{align*}
where $\pi_{m,i,c}$ is the probability that model $m$ answers item $i$ correctly
in condition $c$ ($c=1$ for in-language, $c=0$ for English; each item contributes
both), and $\logit p = \log\frac{p}{1-p}$. The fixed effects are the intercept
$\beta_0$, the log-size slope $\beta_1$, the in-language offset $\beta_2$, and the
size--condition interaction $\beta_3$; $\lsize{m}$ is the centered log (active)
parameter count of model $m$. The random effects are a correlated family
intercept and slope $(b_{0,\famof{m}},\,b_{1,\famof{m}})$, a language intercept
$u_{\langof{i}}$, and an item intercept $v_i$, each Gaussian with mean zero.

\section{Potential Risks}
\label{app:risks}

\bench{} is designed to improve regional-awareness in multilingual evaluation. The resource and its analyses still carry risks, which we flag here for responsible use.

\paragraph{Overclaiming regional competence.} \bench{} does not support comparing model knowledge across languages or countries, because the data is not parallel: each language--country pair draws on different exams, domains, and difficulty levels. We are careful to avoid such comparisons in our own framing, and we ask users of the benchmark to do the same. High accuracy on \bench{} also does not show that a model is safe to deploy in the regions it covers: the benchmark uses a multiple-choice format, and a model that answers items from academic, professional or licensing exams well may still produce unsafe or incorrect output in open-ended use. % We release \bench{} as a diagnostic tool, not a certificate of regional readiness, and caution against using leaderboard scores to justify deployment in high-stakes settings.

\paragraph{Simplification of regional views.} Treating one answer as correct for an entire region can present a single authority's view as fact. Some \bench{} items include historical interpretations, religious or political positions, and disputed events. Scoring a model against one fixed option rewards that view and penalizes minority or contested perspectives. Used to train or select models, the benchmark could propagate these views and present contested topics as settled to end users. We frame each question as a test of whether a model matches a stated regional context, not a judgment about which view is right.

\paragraph{Benchmark contamination and overfitting.} Public release exposes \bench{} to two risks. First, its items may enter future pretraining corpora, inflating scores without real gains in regional knowledge. Second, the released per-country knowledge-dark analysis enables targeted optimization for weak regions rather than broad improvement.

%We additionally examine whether scale reduces cross-language unevenness by regressing each family’s in-language accuracy spread across languages on log model size.

\begin{table*}[t]
\centering
\small
\resizebox{0.7\textwidth}{!}{%
\begin{tabular}{lcccc}
\toprule
Setting & Global winner & Mean $\rho$ & Median $\rho$ & Winner changes \\
\midrule
Original-language & Gemini 3.5 Flash & 0.859 & 0.952 & 39.6\% (36/91) \\
English-prompt    & GPT-5.5 (low)    & 0.888 & 0.970 & 69.2\% (63/91) \\
\bottomrule
\end{tabular}%
}
\caption{Stability of global model rankings under language-specific evaluation. $\rho$ denotes Spearman rank correlation between the global leaderboard and each
language-specific leaderboard. Winner changes indicate languages where the global winner is not the local winner after bootstrap filtering.}
\label{tab:ranking_stability}
\end{table*}

\begin{table*}[h]
\centering
\small
\resizebox{0.8\textwidth}{!}{%
\begin{tabular}{l r l r r c r}
\toprule
Language & $n$ & Local winner & Local acc. & Global acc. & Winner changed & $\rho$ \\
\midrule
Oromo    &   993 & Gemma 3-4B          & 26.99 & 23.26 & Yes & 0.007 \\
Tangale  &   273 & Mistral 7B-instruct & 34.07 & 25.27 & Yes & 0.166 \\
Maithili &   245 & Gemma 3-12B         & 47.06 & 37.14 & Yes & 0.380 \\
Jju      &   599 & Mistral 7B-instruct & 34.39 & 29.55 & Yes & 0.390 \\
Ewe      &   350 & Gemini 3.5 Flash    & 45.14 & 45.14 & No  & 0.458 \\
Ga       & 1,837 & Gemini 3.5 Flash    & 46.33 & 46.33 & No  & 0.497 \\
Dangme   &   569 & Gemini 3.5 Flash    & 46.05 & 46.05 & No  & 0.504 \\
Galician &   350 & Gemini 3.5 Flash    & 31.71 & 31.71 & No  & 0.556 \\
Dagbani  & 1,160 & Gemini 3.5 Flash    & 56.03 & 56.03 & No  & 0.609 \\
Idoma    &   642 & GPT-5.5       & 47.04 & 42.68 & Yes & 0.626 \\
\bottomrule
\end{tabular}%
}
\caption{The ten languages for which the model ranking of performance in that language correlates least with the global model ranking. Evaluated in the in-language
setting.}
\label{tab:rank_stability_worst10}
\end{table*}

\begin{table*}[h]
\centering
\small
\resizebox{0.9\textwidth}{!}{%
\begin{tabular}{l r l r r c r}
\toprule
Language & $n$ & Local winner & Local acc. & Global acc. & Winner changed & $\rho$ \\
\midrule
Kannada     &    541 & GPT-5.5    & 78.37 & 78.19 & Yes & 0.988 \\
Azerbaijani &  6,143 & GPT-5.5    & 83.93 & 82.99 & Yes & 0.986 \\
Dutch       &  1,446 & GPT-5.5    & 90.53 & 90.32 & Yes & 0.984 \\
Telugu      & 11,237 & Gemini 3.5 Flash & 83.50 & 83.50 & No  & 0.984 \\
Turkish     &  2,765 & GPT-5.5    & 94.79 & 94.32 & Yes & 0.983 \\
Indonesian  & 11,045 & Gemini 3.5 Flash & 87.23 & 87.23 & No  & 0.982 \\
Hebrew      &  2,282 & Gemini 3.5 Flash & 79.84 & 79.84 & No  & 0.981 \\
Gujarati    &    321 & Gemini 3.5 Flash & 92.21 & 92.21 & No  & 0.980 \\
Malayalam   &    886 & Gemini 3.5 Flash & 86.79 & 86.79 & No  & 0.980 \\
Nepali      &  1,316 & Gemini 3.5 Flash & 90.50 & 90.50 & No  & 0.980 \\
\bottomrule
\end{tabular}%
}
\caption{The ten languages for which the model ranking of performance in that language correlates most with the global model ranking. Evaluated in the in-language setting.}
\label{tab:rank_stability_best10}
\end{table*}

\clearpage
\begin{table*}[h]
\centering
\small
\resizebox{0.65\textwidth}{!}{%
\begin{tabular}{lcc}
\toprule
Coefficient & (A) item RE & (B) no item RE \\
\midrule
$\beta_0$ intercept & 0.168 [-0.112, 0.425] & 0.182 [-0.097, 0.437] \\
$\beta_1$ $\log$size & 0.430 [0.138, 0.727] & 0.336 [0.062, 0.619] \\
$\beta_2$ in-lang & -0.371 [-0.616, -0.141] & -0.285 [-0.515, -0.064] \\
$\beta_3$ $\log$size$\times$in-lang & 0.079 [-0.214, 0.447] & 0.058 [-0.272, 0.383] \\
\bottomrule
\end{tabular}%
}
\caption{Logistic mixed-effects fit on region-agnostic items. Outcome is item correctness; $\log$size is centered log parameter count, \emph{cond}$=1$ for in-language (vs.\ English). The coefficient of interest is $\beta_3$: the change in the English-vs-in-language gap as models scale ($\beta_3>0$ means larger models close the gap). Both columns include a per-family random slope on $\log$size and a language random intercept; (A) adds an item random intercept.}
\label{tab:rq6_glmm}
\end{table*}

\begin{table*}[h]
\centering
\small
\resizebox{0.45\textwidth}{!}{%
\begin{tabular}{lrlr}
\toprule
Family & \# & Sizes (B) & $\beta_1^{f(m)}$ \\
\midrule
Apertus & 2 & 8, 70 & 0.079 \\
Aya Expanse & 2 & 8, 32 & 0.284  \\
Gemma & 6 & 0.27, 1, 4, 7, 12, 27 & 0.567 \\
Llama & 2 & 8, 70 & 0.340\\
Mistral & 2 & 7, 24 & 0.415  \\
Qwen & 8 & 0.6, 1.7, 4, 7, 8, 14, 32 & 0.315 \\
% Tiny Aya & 4 & 3.35 & 0.405  \\
\bottomrule
\bottomrule
\end{tabular}%
}
\caption{Per-family scaling slopes from the mixed-effects regression. $\beta_1^{f(m)}$ is the family-specific slope for family $f(m)$ on log parameter count ($\beta_1$ plus the family random-slope deviation), reflecting how quickly accuracy on region-agnostic items grows with size within each family. The regression is fit over a broader set of checkpoints than those reported in Table~\ref{tab:results}: we include smaller variants (\eg, sub-1B Qwen and Gemma checkpoints) that are excluded from the main results table, which is restricted to models in a competitive accuracy range, but that contribute the within-family size variation needed to estimate $\beta_1^{f(m)}$. Tiny Aya is omitted from this analysis because all four checkpoints share parameter count 3.35B, so the within-family slope is not identified.}
\label{tab:rq6_family}
\end{table*}

\begin{table*}[h]
\centering
\small
\adjustbox{max width=0.70\textwidth}{%
\begin{tabular}{lrrrrrr}
\toprule
Country & \#nat & Dark$_{\mathrm{nat}}$\% & At\% & Below\% & Dark$_{\mathrm{spec}}$\% & $\Delta$pp \\
\midrule
Qatar & 259 & \textbf{37.5} & 4.6 & \textbf{32.8} & 38.5 & +1.1 \\
Israel & 117 & 36.8 & 18.8 & 17.9 & 10.3 & -26.5 \\
Ghana & 5708 & 34.2 & 17.6 & 16.6 & 34.1 & -0.1 \\
% Belgium & 9 & 33.3 & 22.2 & 11.1 & 19.0 & -14.3 \\
Hungary & 94 & 30.9 & 13.8 & 17.0 & 22.9 & -8.0 \\
Nigeria & 8406 & 30.0 & 14.1 & 15.9 & 30.0 & +0.0 \\
Ethiopia & 2013 & 25.3 & 9.2 & 16.1 & 25.0 & -0.3 \\
% Argentina & 25 & 24.0 & 8.0 & 16.0 & 11.8 & -12.2 \\
Venezuela & 39 & 20.5 & 5.1 & 15.4 & 19.5 & -1.0 \\
Egypt & 619 & 19.5 & 10.0 & 9.5 & 19.3 & -0.3 \\
Rwanda & 67 & 19.4 & 9.0 & 10.4 & 22.1 & +2.7 \\
Belarus & 404 & 17.8 & 13.9 & 4.0 & 17.9 & +0.1 \\
Paraguay & 2897 & 17.4 & 11.0 & 6.4 & 16.6 & -0.8 \\
Ukraine & 92 & 17.4 & 7.6 & 9.8 & 6.2 & -11.2 \\
Iran & 7175 & 16.8 & 10.5 & 6.3 & 17.5 & +0.7 \\
Finland & 92 & 16.3 & 8.7 & 7.6 & 16.6 & +0.3 \\
Costa Rica & 80 & 16.2 & 3.8 & 12.5 & 15.1 & -1.2 \\
Azerbaijan & 1145 & 16.2 & 8.0 & 8.1 & 16.4 & +0.3 \\
Kenya & 2189 & 15.2 & 6.5 & 8.7 & 15.4 & +0.2 \\
Armenia & 580 & 15.0 & 9.5 & 5.5 & 15.3 & +0.3 \\

\bottomrule
\end{tabular}%
}
\caption{Candidate knowledge-dark regions. An item is marked \emph{knowledge-dark} when the 23-model population performs at or below chance ($\gamma=0.25$) in both English and in-language. 
We report the per-country dark fraction over regional-native (\textit{Region-anchored} and \textit{Region-embedded}) items, decomposed into at-chance and significantly below-chance cases, and compare it to the broader region-specific set, \ie, including \textit{Region-conditional} items as well. $\Delta$pp denotes the change in dark fraction after adding region-conditional items. Countries are sorted by Dark$_{\mathrm{nat}}$; overall dark fractions are 13.6\% for regional-native and 13.4\% for region-specific items.}
\label{tab:knowledge-dark}
\end{table*}

\begin{table*}[t]
\centering
\small
\resizebox{\textwidth}{!}{%
\begin{tabular}{lrrrrrrrrrrr}
\toprule
Model & \multicolumn{2}{c}{\makecell{Accuracy on region-\\specific items (\%)}} & \multicolumn{2}{c}{\makecell{Correct in one\\language only (\%)}} & \multicolumn{3}{c}{\makecell{(model, language)\\combinations}} & \multicolumn{4}{c}{\makecell{Correct in the first language,\\wrong in the second (\%)}} \\
\cmidrule(lr){2-3} \cmidrule(lr){4-5} \cmidrule(lr){6-8} \cmidrule(lr){9-12}
 & IL & EN & IL only & EN only & IL larger & EN larger & Tested & IL$\to$EN & EN$\to$IL$_{\mathrm{spec}}$ & EN$\to$IL$_{\mathrm{agn}}$ & $\Delta$ \\
\midrule
Claude Opus 4.7 & 72.9 & 68.0 & 11.6 & 6.6 & 49 & 2 & 89 & 20.1 & 14.3 & 13.3 & 0.9 \\
Gemini 3.5 Flash & 78.9 & 69.2 & 13.6 & 3.8 & 60 & 0 & 89 & 20.8 & 9.3 & 10.5 & -1.2 \\
GPT-5.4-mini & 67.9 & 63.6 & 12.3 & 8.0 & 44 & 0 & 89 & 22.8 & 17.5 & 18.4 & -0.9 \\
GPT-5.5 & 76.8 & 69.6 & 11.6 & 4.4 & 55 & 0 & 88 & 19.2 & 10.7 & 12.1 & -1.5 \\
\midrule
\emph{Frontier} (4 models) & -- & -- & -- & -- & 208 & 2 & 355 & -- & -- & -- & -- \\
\midrule
Apertus-70B & 38.3 & 40.8 & 14.6 & 17.2 & 14 & 38 & 90 & 43.8 & 47.1 & 46.4 & 0.7 \\
Apertus-70B-think & 39.6 & 39.7 & 16.8 & 16.9 & 15 & 23 & 91 & 47.4 & 48.1 & 46.6 & 1.4 \\
Apertus-8B & 38.1 & 41.4 & 14.2 & 17.5 & 6 & 29 & 90 & 42.5 & 46.9 & 48.0 & -1.2 \\
Aya Expanse-32B & 47.4 & 52.1 & 13.5 & 18.3 & 8 & 42 & 90 & 34.7 & 39.9 & 40.3 & -0.4 \\
Aya Expanse-8B & 38.3 & 47.2 & 13.0 & 21.8 & 2 & 53 & 91 & 38.7 & 49.4 & 50.7 & -1.3 \\
Gemma 3-12B & 52.1 & 53.5 & 12.9 & 14.3 & 10 & 17 & 90 & 30.7 & 31.6 & 28.6 & 3.0 \\
Gemma 3-27B & 56.2 & 55.6 & 12.6 & 12.0 & 16 & 12 & 90 & 27.2 & 26.5 & 23.7 & 2.8 \\
Gemma 3-4B & 42.4 & 46.3 & 14.6 & 18.5 & 1 & 42 & 90 & 37.9 & 43.2 & 38.7 & 4.5 \\
Gemma 4-31B & 64.9 & 61.1 & 11.7 & 7.8 & 39 & 0 & 89 & 23.5 & 18.6 & 17.6 & 1.1 \\
Llama 3.1-70B & 53.6 & 56.2 & 12.7 & 15.3 & 9 & 29 & 89 & 30.2 & 33.3 & 34.7 & -1.3 \\
Llama 3.1-8B & 32.9 & 43.9 & 13.0 & 24.0 & 0 & 66 & 90 & 44.2 & 57.5 & 54.2 & 3.3 \\
Mistral-7B & 35.1 & 45.6 & 12.7 & 23.1 & 1 & 65 & 90 & 39.2 & 52.3 & 52.0 & 0.3 \\
Qwen 3-14B & 52.8 & 55.9 & 11.8 & 15.0 & 9 & 28 & 90 & 28.8 & 32.4 & 28.0 & 4.4 \\
Qwen 3-32B & 53.6 & 56.6 & 12.2 & 15.1 & 11 & 27 & 90 & 29.3 & 32.7 & 26.9 & \textbf{5.8} \\
Qwen 3.5-27B & 54.3 & 54.5 & 11.0 & 11.2 & 19 & 15 & 89 & 27.0 & 27.7 & 23.7 & 3.9 \\
Qwen 3-8B & 50.0 & 53.7 & 11.9 & 15.6 & 10 & 39 & 89 & 30.2 & 34.8 & 29.8 & 5.0 \\
Tiny Aya Earth & 34.1 & 38.9 & 15.6 & 20.3 & 3 & 37 & 91 & 48.8 & 54.1 & 51.0 & 3.1 \\
Tiny Aya Fire & 34.8 & 41.4 & 13.7 & 20.3 & 1 & 39 & 90 & 43.1 & 50.9 & 48.9 & 1.9 \\
Tiny Aya Global & 33.2 & 38.7 & 15.2 & 20.6 & 5 & 36 & 91 & \textbf{49.0} & 55.8 & 53.8 & 2.0 \\
Tiny Aya Water & 34.8 & 40.6 & 14.6 & 20.4 & 2 & 37 & 90 & 45.9 & 52.4 & 48.4 & 4.0 \\
\midrule
\emph{Open-weight} (20 models) & -- & -- & -- & -- & 181 & 674 & 1800 & -- & -- & -- & -- \\
\bottomrule
\end{tabular}%
}
\caption{Cross-lingual accuracy asymmetries on region-specific items (\textit{Region-conditional}, \textit{Region-anchored} and \textit{Region-embedded}), macro-averaged over languages. IL is the question in its original language, EN the English translation. The first group gives accuracy on region-specific items in each of the two conditions. \emph{IL only} and \emph{EN only} are the shares of region-specific items answered correctly in one of the two languages and incorrectly in the other. An exact two-sided McNemar test is applied to each combination of one model and one language holding at least one such item. \emph{Tested} counts those combinations; \emph{IL larger} counts the ones where the test rejects at $\alpha=0.05$ and the items answered correctly only in the original language outnumber those answered correctly only in English; \emph{EN larger} counts the ones where it rejects and that inequality is reversed. In the last group the arrow reads correct in the first language, wrong in the second: IL$\to$EN is $\Pr(y^{\mathrm{eng}}{=}0\mid y^{\mathrm{lang}}{=}1)$, and EN$\to$IL is $\Pr(y^{\mathrm{lang}}{=}0\mid y^{\mathrm{eng}}{=}1)$ on region-specific items (written $D_{\mathrm{spec}}$ in the text) and on region-agnostic items ($D_{\mathrm{agn}}$). $\Delta=D_{\mathrm{spec}}-D_{\mathrm{agn}}$ is the part of the drop not attributable to general difficulty with the original language. Both EN$\to$IL columns and $\Delta$ are averaged over the languages where both are defined, so $\Delta$ equals the difference of the two columns beside it. Subtotal rows sum the three count columns; rate columns are left blank there. Of 355 frontier combinations the test rejects for 210, of which 208 favour the original language; of 1800 open-weight combinations it rejects for 855, of which 674 favour English.}
\label{tab:cross-lingual}
\end{table*}

\begin{table*}[t]
\centering
\small
\resizebox{0.8\textwidth}{!}{%
\begin{tabular}{llrrrrr}
\toprule
Language & Country & $n$ & Gap & Lang. Mean Gap & Gap Residual & 95\% CI \\
\midrule
Arabic & Egypt     & 659  & +0.1 & -1.2 & +1.3 & [ +0.5, +2.2] \\
Arabic & Palestine & 134  & -0.3 & -1.2 & +0.9 & [ -0.5, +2.4] \\
Arabic & Jordan    & 2471 & -0.4 & -1.2 & +0.8 & [ +0.2, +1.5] \\
Arabic & Qatar     & 275  & -0.6 & -1.2 & +0.6 & [ -0.5, +1.5] \\
Arabic & Morocco   & 193  & -2.6 & -1.2 & -1.4 & [ -2.7, -0.1] \\
Arabic & UAE       & 77   & -3.5 & -1.2 & -2.3 & [ -4.2, -0.4] \\
\midrule
Chinese & China     & 5361 & +3.6 & +1.2 & +2.4 & [ +1.7, +3.0] \\
Chinese & Hong Kong & 263  & -1.2 & +1.2 & -2.4 & [ -3.0, -1.7] \\
\midrule
Portuguese & Brazil   & 176 & +1.4 & +1.0 & +0.4 & [ -0.4, +1.3] \\
Portuguese & Portugal & 253 & +0.5 & +1.0 & -0.4 & [ -1.3, +0.4] \\
\midrule
Spanish & Costa Rica           & 106  & +2.3 & +0.2 & +2.2 & [ +0.7, +3.8] \\
Spanish & Puerto Rico          & 45   & +1.5 & +0.2 & +1.4 & [ -1.4, +4.0] \\
Spanish & Cuba                 & 380  & +1.4 & +0.2 & +1.2 & [ +0.3, +2.2] \\
Spanish & Paraguay             & 211  & +1.3 & +0.2 & +1.1 & [ +0.0, +2.4] \\
Spanish & Bolivia              & 834  & +1.1 & +0.2 & +0.9 & [ +0.2, +1.6] \\
Spanish & Colombia             & 777  & +0.8 & +0.2 & +0.6 & [ -0.1, +1.4] \\
Spanish & El Salvador          & 268  & +0.7 & +0.2 & +0.5 & [ -0.6, +1.6] \\
Spanish & M\'exico             & 861  & +0.5 & +0.2 & +0.4 & [ -0.3, +1.1] \\
Spanish & Ecuador              & 1522 & +0.3 & +0.2 & +0.2 & [ -0.4, +0.8] \\
Spanish & Espa\~na             & 3961 & +0.2 & +0.2 & +0.0 & [ -0.5, +0.5] \\
Spanish & Chile                & 456  & -0.1 & +0.2 & -0.3 & [ -1.1, +0.6] \\
Spanish & Argentina            & 228  & -0.2 & +0.2 & -0.3 & [ -1.6, +0.9] \\
Spanish & Per\'u               & 1154 & -0.4 & +0.2 & -0.6 & [ -1.3, +0.1] \\
Spanish & Uruguay              & 85   & -1.6 & +0.2 & -1.8 & [ -3.4, -0.1] \\
Spanish & Venezuela            & 41   & -1.8 & +0.2 & -1.9 & [ -4.5, +0.6] \\
Spanish & Panam\'a             & 40   & -4.4 & +0.2 & -4.6 & [ -7.0, -2.0] \\
\bottomrule
\end{tabular}%
}
\caption{Within-language residuals of the IL--EN gap for pluricentric languages on region-specific items (all categories except Region-agnostic). Gap is defined as $\mathrm{Acc}_{\mathrm{IL}}-\mathrm{Acc}_{\mathrm{EN}}$, and Lang. Mean Gap is the average gap across countries sharing the same language. Gap Residual subtracts this language-level mean from each country gap, measuring where the original-language advantage is larger or smaller than the language average. Countries with fewer than 30 non-agnostic samples are discarded. Bootstrap confidence intervals (CI) are computed by resampling items within each language--country cell.}
\label{tab:pluricentric-gap-residuals}
\end{table*}

\end{document}